\documentclass{nle}

\usepackage{multirow}
\usepackage{color}
\usepackage{graphicx}
\usepackage{url}
\usepackage{subfigure} 
\newcounter{tbsnr}
\newenvironment{tbs}
{\addtocounter{tbsnr}{1}\par\bigskip\noindent\fbox{\thetbsnr}
\hspace*{\fill}\begin{minipage}{7cm}\tt}
{\end{minipage}\hspace*{\fill}\bigskip}
\newcommand{\tb}[1]{\begin{tbs}{#1}\end{tbs}}

\newcommand{\cut}[1]{}

\newcommand{\rb}[1]{\textbf{RB: #1}}

\title[Pay attention to those sets!]
      {Pay attention to those sets!\\ Learning quantification from images}
\author[Sorodoc, Pezzelle, Herbelot, Dimiccoli, Bernardi]
       {Sorodoc, I.,\textsuperscript{1} Pezzelle, S.,\textsuperscript{2} Herbelot, A.,\textsuperscript{3} Dimiccoli, M.,\textsuperscript{4} Bernardi, R.\textsuperscript{5}\\
        \textsuperscript{1,2,3}CIMeC - Center for Mind/Brain Sciences, \textsuperscript{5}CIMeC/DISI, University of Trento\\
        \textsuperscript{4}CVC - Computer Vision Center, University of Barcelona\\
        \textsuperscript{1,2,3,5}\{firstname.lastname\}@unitn.it\\
        \textsuperscript{4}\{mariella.dimiccoli\}@cvc.uab.es}

\begin{document}

\label{firstpage}
\maketitle

\begin{abstract}

Major advances have recently been made in merging language and vision representations. But most tasks considered so far have confined themselves to the processing of objects and lexicalised relations amongst objects (\textit{content words}). We know, however, that humans (even pre-school children) can abstract over raw data to perform certain types of higher-level reasoning, expressed in natural language by \textit{function words}. A case in point is given by their ability to learn quantifiers, i.e. expressions like \emph{few}, \emph{some} and \emph{all}.

From formal semantics and cognitive linguistics, we know that quantifiers are relations over sets which, as a simplification, we can see as proportions. For instance, in \textit{most fish are red}, \emph{most} encodes the proportion  of fish which are red fish.  In this paper, we study how well current language and vision strategies model such relations.  We show that state-of-the-art attention mechanisms coupled with a traditional linguistic formalisation of quantifiers gives best performance on the task.

Additionally, we provide insights on the role of `gist' representations in quantification.  A `logical' strategy to tackle the task would be to first obtain a numerosity estimation for the two involved sets and then compare their cardinalities. We however argue that precisely identifying the composition of the sets is not only beyond current state-of-the-art models but perhaps even detrimental to a task that is most efficiently performed by refining the \textit{approximate} numerosity estimator of the system.

\end{abstract}

\section{Introduction}
\label{sec:intro}

Natural language sentences are built from complex interactions between
\textit{content} words (e.g., nouns, verbs) and \textit{function}
words (e.g., quantifiers, coordination). A well-founded,
broad-coverage semantics should therefore jointly model lexical items
and functional operators~\cite{bole:form16}. Computational work on
language and vision, however, has so far mostly focused on the
lexicon, and topical representations of text fragments. One strand of
work concentrates on content word representations, and nouns in
particular (see for example~\cite{Anderson2013,Lazaridou2015}), whilst
another is interested in approximate sentence representation, as in
the Image Captioning (IC) and the Visual Question Answering tasks
(VQA) (e.g.,~\cite{Hodosh:etal:2013,xu:show15,anto:vqa15,goya:maki16}).  Our work
aims at filling the gap on the functional side of language, by
exploring the performance of language and vision models on a
particular logical phenomenon: \textbf{quantification}.

\begin{figure}[t]
\begin{center}
\begin{tabular}{ccc}

(a)  \includegraphics[width=0.25\linewidth]{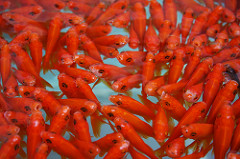}&
(b) \includegraphics[width=0.22\linewidth]{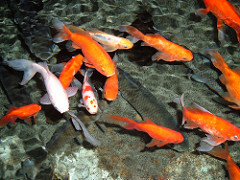} &
(c) \includegraphics[width=0.25\linewidth]{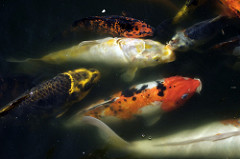} \\
(d)  \includegraphics[width=0.25\linewidth]{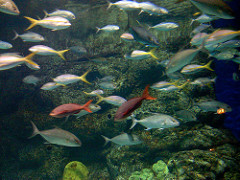} &
(e) \includegraphics[width=0.25\linewidth]{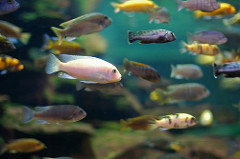} &

\end{tabular}
\end{center}
\caption{Query: \underline{\hspace{5mm}}  fish are red. 
Answers: (a)  All, (b) Most, (c) Some, (d) Few, (e) No.\protect\footnotemark}
\label{fig:differentanswers}
\end{figure}

Quantification has been marginally studied in recent work on language
and vision, in the context of VQA, focusing on
`number questions' that can be answered with cardinals. It has been
found that out-of-the-shelf state-of-the-art (SoA) systems perform
poorly on the type of questions~\cite{ren:imag15,anto:vqa15} which
requires \emph{exact numerosity estimation}, although recent work
shows that it might be possible to adapt them to the counting
task~\cite{pari:count16}. In this paper, we focus on a complementary
phenomenon by considering quantifiers which involve a) an
\emph{approximate number estimation} mechanism; and b) a
\emph{quantification comparison} step, i.e. the computation of a
proportion between two sets. For instance, given the images in
Figure~\ref{fig:differentanswers}, we want to quantify which
proportion of fish are red fish. This endeavour, as we
  argue below, is not simply an investigation of a different type of
  quantifier. We claim that this specific problem is an interesting
  opportunity to reflect on the way we build neural network
  architectures.

\footnotetext{Pictures: flickr CC-BY. (a) from user Hapal (\url{https://www.flickr.com/photos/hapal/}; (b) from user Brett Vachon (\url{https://www.flickr.com/photos/asdsoupdsa/}); (c) from user Jenn (\url{https://www.flickr.com/photos/happyskrappy/}); (d) from user Micha Hanson (\url{https://www.flickr.com/photos/denverhansons/}); (e) from user Jeff Kubina (\url{https://www.flickr.com/photos/kubina}).}


At the linguistic level, formal semanticists have extensively studied
these expressions (so-called `generalised quantifiers') and described
them as relations between a \emph{restrictor} (e.g., \textit{fish}),
which selects a set of target objects in a state-of-affairs, and a
\emph{scope} (e.g., \textit{red}) which selects the subset of the
target set which satisfies a certain property.  Alternatively, they
can be seen as proportions between the selected sets, e.g., $|
\mbox{red fish} | / |\mbox{fish} |$.

This proportional property of generalised quantifiers necessitates an
operation at a level of abstraction which, we think, is interestingly
different from the level of shallow reasoning needed to process
content words and simple cardinals.  The intuition behind our
conjecture can be explained by considering the following. Let's assume
that we want to find the correct quantifier for a particular
concept-feature pair (e.g., fish-red), given a specific image (see
Fig~\ref{fig:differentanswers}, where the task is to return which
proportion of fish are red). We want the network to learn that certain
quantifiers correspond to certain set configurations: given sets $A$
and $B$, if $A \cap B$ is nearly entirely contained in $A$, then it is
true that \textit{most As are Bs}; if the overlap is less, then
\textit{few} or \textit{some As are Bs}. There is here a correlation
to be learnt between different set configurations and particular
quantifiers, but those configurations are \textit{abstractions} over
the raw linguistic and visual data: when the set comparison takes
place, it is irrelevant whether $A$s are fish or ice cream scoops, or
indeed, how many $A$s exactly were observed.  In fact, as we argue
below, trying to integrate this information in the quantification
decision may be detrimental to the system.

Quantifiers are operators which can be applied to any set, regardless
of its composition and whether it matches statistics observed at the
category level. So attempting to use category-level information (e.g.,
generally speaking, 20\% of all fish are red) will result in failure
to generalise to randomly sampled subsets of small
cardinality. Fig~\ref{fig:differentanswers} illustrates the point,
where knowledge of fish or redness is not enough for the predictive
power of the system.\footnote{We note that in some special cases,
  there \textit{is} a correlation between certain concept-property
  pairs and their quantification: in particular, definitional
  properties correspond to universal quantification (for instance,
  \textit{triangle} and \textit{three-sided} can always be quantified
  with \textit{all}). However, those special cases only apply to
  universal quantification.} Similarly, the amount of overlap between
two sets can be associated with particular quantifiers regardless of
the cardinality of those two sets, what matter is their proportion. So
an ideal model will learn to abstract over cardinality information too.

\cut{
\begin{figure}
\begin{center}
\begin{tabular}{c}
\includegraphics[width=0.70\linewidth]{./images/abstraction} 
\end{tabular}
\caption{Abstract representation}\label{fig:abstract}
\end{center}
\end{figure}
}

The most straightforward and efficient strategy to learn to quantify
could be to divide the task into two subtasks: learning to generalize
the correlation (a) from raw data to their abstract representation and
(b) from the latter to quantifiers.  The high results obtained
in~\cite{soro:look16}, who have trained NNs to quantify over synthetic
scenarios of coloured dots, suggest that NNs should be able to learn
the second subtask quite easily. In this paper, we study how far
current strategies to integrate the language and vision modalities are
suitable when put to work on the full task, involving quantification over real-life images.  We revisit some state-of-the-art VQA models, considering some of the NN features which may affect how the model deals with this high-level process. In particular, we focus
on a) the role of sequential processing in both modalities and the b)
attention mechanisms, within and across modalities, which are at the
core of many state-of-the-art systems.

We show that, as in the case of content words, attention
mechanisms help obtaining a more salient representation of the
linguistic and visual input, useful for the processing of quantifiers. As observed above, in contrast with content words, functional operators act over sets.  An approximate, visually-grounded representation of such sets can be obtained by exploiting the logical
structure of the linguistic query, combined with attention. 
More concretely, we show that when
dealing with quantifiers instead of computing the composed
representation of the linguistic query and then use it to attend the
image, it is better to reach a multimodal composition by using the
linguistic representation of the restrictor to guide the visual
representation of the scenarios, and then the latter to guide the
composition of the linguistic representation of the restrictor with
the linguistic representation of the scope. Our results highlight that
using the output of an LSTM on the language side to attend to the relevant parts of the image is less successful than this attention mechanism.

Additionally, we provide insights on the role the image gist representation, built by attention models, has in the quantification task.  A `logical' strategy to tackle the quantification task would be to first obtain the numerosity estimation of the two involved sets and then compare their quantities. This method could be implemented by aiming to extract a fully abstract representation of the sets in the raw data. We however argue that, given the inherent difficulty in identifying objects, and even more, properties, an approximate set representation in the form of a visual gist may be a more efficient and cognitively plausible strategy.

Finally, we should mention that our work touches on the current
debate of balancing datasets of natural images.  \cite{zhou:simp15},
for example, have demonstrated that a simple bag-of-word baseline,
that concatenates visual and textual inputs, can achieve very decent
overall performance on the VQA task. That is, the performance of the
model is due to the excellent ability of the network to encode certain
types of correlations, either within or across modalities. Part of
these results might be due to the language prior that has been
discovered in the VQA dataset~\cite{zhan:yin16,girs:clev16} and that
has been addressed by either using abstract scenes or by carefully
building a dataset of very similar natural images corresponding to
different answers~\cite{goya:maki16}. The quantification dataset we
propose in \S\ref{sec:data} of this paper follows this intuition,
making sure that the entity sets that the system is required to
quantify over do not exhibit unwanted regularities.

\section{Related Work}
\label{sec:related_work}


\paragraph{Computational models of quantifiers}

The problem of algorithmically describing logical quantifiers was first addressed by~\cite{vanb:essa86} using automata. Following these first efforts, a lot of work has been done in computational formal semantics to model quantifiers in language (see e.g.~\cite{szab:quant10,hand:keen} for an overview). Recently, distributional semantics has turned to the problem, with~\cite{Baroni2012} demonstrating that some entailment relations hold between quantifier vectors obtained from large corpora, and~\cite{Herbelot2015} mapping a distributional vector space to a formal space from which the quantification of a concept-property pair can be predicted. This line of work, however, only considers the linguistic modality, without attention to vision.

In parallel to the formal linguistic models, psycholinguistics has studied function
words from a statistical perspective using NN architectures. At the end of the 90s,  \cite{deha:deve93} showed how approximate numerosity could be extracted from visual input without serial counting, bringing computational evidence to the psycholinguistic observation that infants develop numerosity abilities before being able to count. Of particular interest to us, \cite{raja:groun05} aimed at grounding linguistic quantifiers in perception. The quantifiers studied were \textit{a few, few, several, many} and \textit{lots}, and the system was trained on human annotations of images consisting of white and stripy fish.  Given an image, the model had to predict which proportion of fish was stripy, using the given quantifiers.  The authors showed that both spacing and the number of objects played a role in the prediction.  

These studies were touching upon an interesting research avenue, but the NN models available at the time were not powerful enough for a full investigation. In the meantime, interesting progress on modelling the acquisition of quantifiers in a Bayesian probabilistic framework has been reported in~\cite{pian:mode12,pian:lear11}. More recently, NNs have been shown to perform well in tasks related to quantification,
from counting to simulating the Approximate Number Sense (ANS). Segu\'{i} et al.~\shortcite{Segui2015}, for instance, explore the task of counting occurrences of an object in an image using convolutional NNs, and demonstrate that object identification can be learnt as a surrogate of counting.  Stoianov and Zorzi \shortcite{Stoianov2012} show that the ANS emerges as a statistical property of images in deep networks that learn a hierarchical generative model of visual input. Very interesting models have also been
proposed by \cite{pari:count16}, who focus on the issue of counting everyday objects in visual scenes, using subitising strategies observed in humans. Similarly focusing on the subitising process, \cite{zhan:sali15} address the issue of salient object detection and show how CNN models can discriminate between images with 0 to 4+ salient objects. As discussed in~\cite{li:sali14}, the salient object detection task
highly depends on various properties of the images, like the uniformity of the various regions, the complexity of the foreground and background, how close to each other the salient objects are, and how they differ in size. 


The models we present in this paper can be seen as a continuation of previous work on linguistic quantifiers. As in \cite{deha:deve93}, the systems we evaluate do not rely on explicit counting, and use the gist of the objects in an image to produce the appropriate quantifier for a given scenario. We also follow~\cite{raja:groun05} in their investigation of `vague' linguistic quantifiers, but we train and evaluate our system on real images rather than toy examples. Unlike them, however, we do not investigate object position in the image and start from their bounding boxes.

To our knowledge, \cite{soro:look16,pezz:bepr16} are the only recent attempt to model non-cardinals in a visual quantification task, using neural networks. \cite{pezz:bepr16} focus on the difference between the acquisition of cardinals and quantifiers, showing they can be modelled by two different operations within the network, and learning one function per  cardinal/quantifier. Our paper can be seen as extending the work of~\cite{soro:look16} by a) augmenting their list of logical quantifiers (\textit{no, some, all}) with proportional ones (\textit{few, most}); b) moving from artificial scenarios with geometric figures to real images; c) most importantly, treating quantifiers as relations between two sets of objects amongst a number of distractors (in contrast, their scenarios only include objects of the same type, e.g. circles, and the task is to quantify over the colour property of those circles).

\paragraph{Datasets with numerosity annotation}


COCO-QA~\cite{ren:expl15} was the first dataset of images associated with number questions. COCO-QA consists of around 123K images extracted from~\cite{lin:micro14}, and 118K questions generated automatically from image descriptions. Number questions are one of the four question categories (together with object, color and location) and make up  7.47\% of the overall questions both in the training and test datasets.  For this category, the authors observe that the evaluated models can sometimes count up to 5 or 6. However, this ability is fairly weak as they do not count correctly when presented with unknown object types. Starting from~\cite{lin:micro14}, \cite{anto:vqa15} built the VQA dataset, aiming to increase the diversity of knowledge and kinds of reasoning needed to provide correct answers. VQA consists of around 200K images 614K questions, and 6M ground truth answers. It contains open-ended, free-form questions and answers provided by humans. The evaluation of SoA models against this dataset confirmed that number questions are
hard to be answered and are those for which a good combined understanding of the language and vision modalities is essential. The difficulty of number questions was further highlighted in~\cite{girs:clev16}, where the authors introduced CLEVR (Compositional Language and Elementary Visual Reasoning diagnostics), a dataset allowing for an in-depth evaluation of current VQA models on various visual reasoning tasks. The reasoning skills they investigated (querying object attributes, counting sets of objects or comparing values, existence questions) are close to the task we propose. They show that state-of-the-art systems perform poorly in situations requiring short-term memory (attribute comparison and integer equality).  

Focusing on the subitising phenomenon, the Salient Object Subitising (SOS) dataset, proposed in~\cite{zhan:sali15}, contains about 14K everyday images annotated with respect to numerosity of salient objects (from 0 to 4+). Images were gathered from various sources (viz. COCO, ImageNet, VOC07, and SUN) and filtered out to create a balanced distribution of images containing obviously salient objects. To eliminate the bias due to unbalanced number distribution (indeed, most of the images contained 0 or 1 salient object), the authors used a cut-and-paste strategy and generated synthetic SOS image data. 

None of the datasets above meets our needs for the quantification task. In SOS images, salient objects are all of the same category and properties are not annotated. Only small numerosities are represented. As for VQA, it does contain annotated objects of different categories but does not provide properties annotation.  Very recently,however,  a new version of COCO-QA, COCO Attribute-QA, has been released. It contains images annotated with both objects (of various categories) and properties~\cite{patterson2016coco}. It consists of 84K images, 180K unique objects (from 29 object categories) and 196 attributes, for a total of 3.5M object-attribute pairs. We take this as our starting point to create a dataset of natural images which can be matched to the range of quantifiers in our study (see \S\ref{sec:data}).


\paragraph{Neural Networks for VQA}

Since the pioneer work by~\cite{mali:amul14,gema:visu15}, many researchers have taken up the VQA challenge. Most of them have based their system on Neural Network models~\cite{gao:arey15,ren:imag15,mali:asky15,ma:lear16} that can learn to perform the given task in an end-to-end fashion.  The first NNs proposed to tackle VQA were based on a combination of global visual feature vectors extracted by a convolutional neural network (CNN), and text feature vectors extracted using a long-short term memory network (LSTM). Various \emph{LSTM-CNN models} have been proposed which differ with regard to the way these two types of features \emph{are combined} (multimodal pooling): by mere concatenation~\cite{zhou:simp15}, or by more complex operations like element-wise multiplication~\cite{anto:vqa15} or multimodal compact bilinear pooling~\cite{fuki:mult16}. Proposals have also been made to use only one architecture. \cite{ren:expl15} use an LSTM to jointly model the image and the question: they treat the image as a word appended to the question, and the image is processed by a CNN model, the output of which is frozen during the training process. More recently, on the opposite site, a convolutional architecture has been used to learn both types of feature and their
interaction~\cite{ma:lear16}. 

Significant progress has been made on the VQA task by the introducing \emph{memory} and \emph{attention} components, taken from other areas of LaVi. \cite{xu:show15}, for instance, introduced an attention-based framework into the problem of image caption generation. Memory Networks (MNs) have been used to tackle tasks involving reasoning on natural language text~\cite{west:memo15,Sukhbaatar:etal:2015new}. A combination of both the memory and attention components have been proposed by e.g.~\cite{kuma:askm16} and recently applied to the VQA challenge in the Dynamic Memory Network (DNM+)~\cite{xion:dyna16} and Stacked Attention Networks (SANs)~\cite{yang:stac16}. \cite{andr:learn16,andre:neur16} further combine the dynamic properties of previous models with the compositionality process of natural language questions via reinforcement learning. 

We build on this previous work by porting insights from the VQA task to our quantification task. In particular, we investigate the role of LSTMs and their combination with CNNs, both as simple concatenation and within stacked attention mechanisms.
In the end, we propose a model that combines formal semantics intruitions about quantifiers (as relations between a restrictor and a scope), and the latest findings of VQA models on attention mechanisms.

\section{Data}
\label{sec:data}

For our task, the required datapoints will be of the form $\langle query, scenario, answer \rangle$. The $answer$ is a quantifier (\textit{no, few, some, most} or \textit{all}). The $query$ is an $\langle$object, property$\rangle$ pair (e.g., $\langle$ dog, black $\rangle$) such that the object and the property correspond to the restrictor and scope of the quantifier, respectively. The $scenario$ is an image containing objects which may or may not be of the type of the restrictor, and may or may not have the property expressed by the scope. We will refer to the objects that have the required property (e.g., black dogs) as
\emph{target} objects.  

We take quantifiers to stand for fixed relations (operationalised as proportions) between the relevant sets of objects: $\frac{|restrictor \cap scope|}{|restrictor|}$. Hence, we take \textit{no} and \textit{all} to be the correct answer for scenarios in which the target objects are 0\% and
100\% of the restrictor set, respectively. To define \textit{few} and \textit{most}, we use prevalence estimates reported by~\cite{khemlani2009generics} for \textit{low-prevalence} and \textit{majority} predications. In particular, we assign \textit{few}
to ratios lower or equal to $17\%$, and \textit{most} to ratios equal or greater than $70\%$. All ratios ranging between these two values
are assigned to \textit{some}. 

\subsection{From COCO ATTRIBUTE to Q-COCO}

\begin{figure}
\begin{center}
  \includegraphics[width=0.9\linewidth]{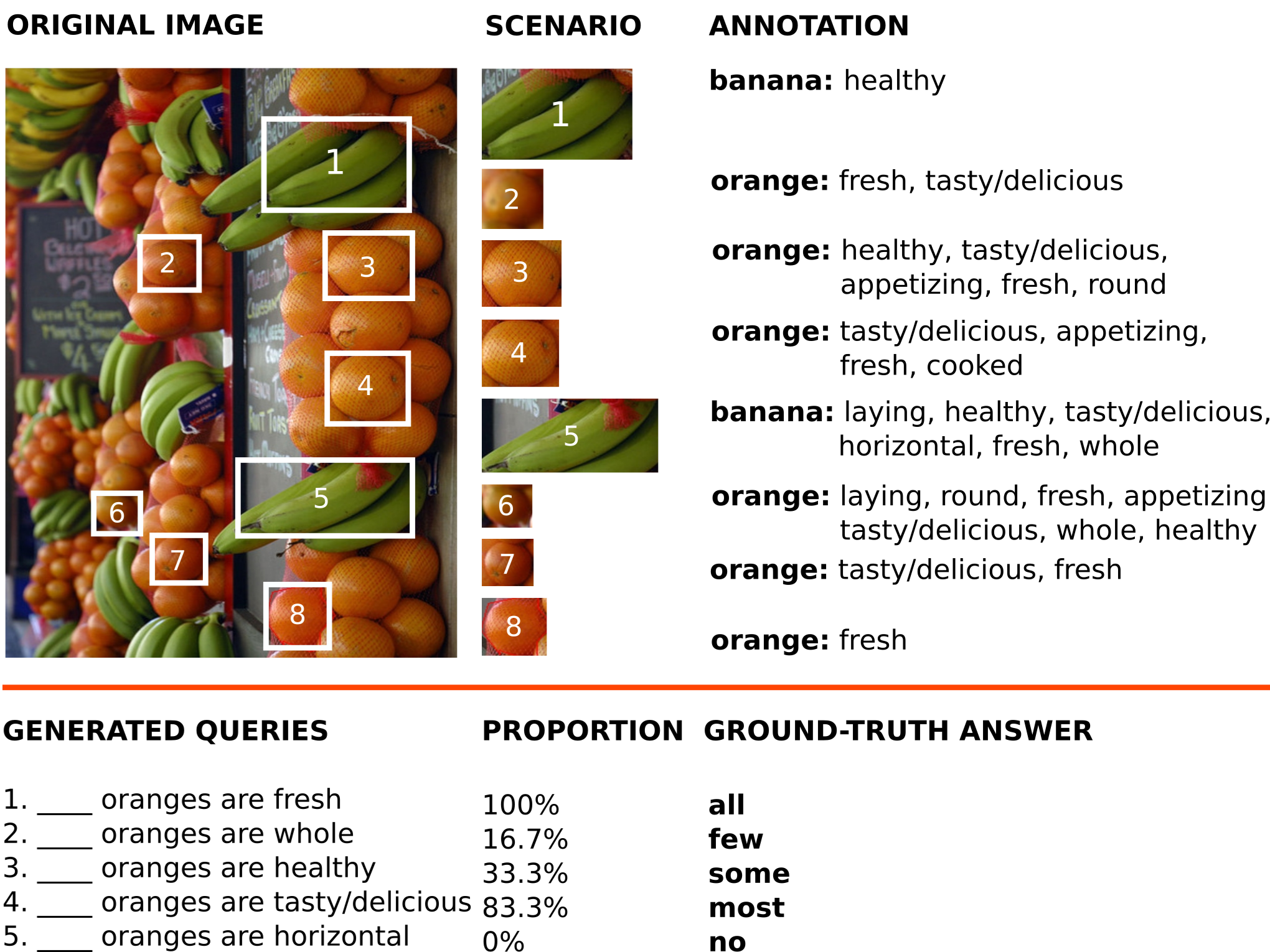}
\end{center}
\caption{Generation of datapoints for the Q-COCO dataset.}
\label{fig:cocodata}
\end{figure}

COCO-Attribute~\cite{patterson2016coco} is a dataset with comprehensive property annotation. It contains 84K image from MS-COCO~\cite{lin:micr14}. Some of the objects are marked with region coordinates/bounding boxes, and their properties (`attributes' in COCO terminology) have been annotated by humans. In total, there are 29 object categories (types of objects), 196 properties and 180K annotated regions, with an average of 19 properties per annotated object.  Not all objects in an image are annotated with respect to properties: only those that are included in the 29 object categories, and for which bounding boxes are provided. Hence, we cannot exploit the full image, but must restrict ourselves to the annotated regions. As illustrated in Figure~\ref{fig:cocodata}, we construct Q-COCO scenarios from this data, following the procedure described below. 

First of all, we filter out all images containing less than 6 annotated objects, thus obtaining 5,203 unique images. This choice is motivated by the fact that 6 is the lowest restrictor cardinality that allows us to have all five quantifiers represented in the data, given the ratios we have assigned to them. To clarify this point, 0 out of 6 objects would be a case of \textit{no}; 1 out of 6 would be \textit{few}; 2, 3 and 4 would be \textit{some}; 5 would be \textit{most}, and 6 would be \textit{all}. Note that if we had used 5 objects instead of 6, \textit{few} would not have been represented. So this constraint is a necessary (though not sufficient) condition to avoid bias due restrictor cardinality. 

Secondly, for each of these 5,203 images, all properties associated with each annotated object are extracted. We compute the overall frequency of each property and, to avoid data sparsity, we retain only properties with frequency $>1000$. That is, if the object `banana' in a given image is originally annotated with 3 properties (e.g., \textit{appetizing, fresh, delicious}), only the most frequent ones are included (e.g., \textit{appetizing, fresh}). This way, we obtain 44 unique properties. Finally, we retain only the images containing at least 6 annotated objects that belong to the same category (e.g., \textit{banana}).

\begin{figure}
\begin{center}
  \includegraphics[width=0.75\linewidth]{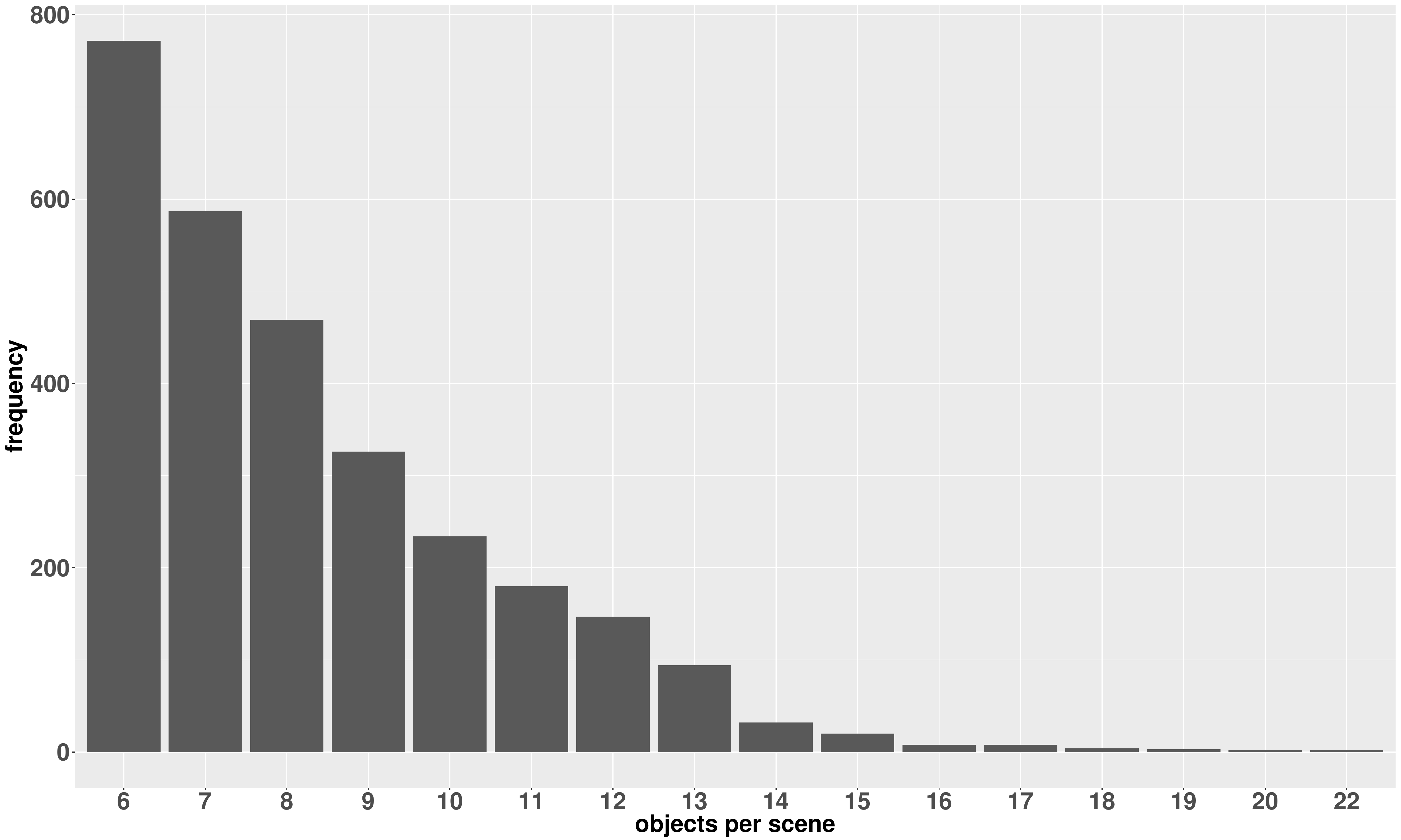}
\end{center}
\caption{Distribution of Q-COCO images against number of annotated objects. Number of annotated objects (X-axis) ranges from 6 to 22.}
\label{fig:distribution}
\end{figure}

As reported in Table~\ref{tab:datasets}, the resulting dataset includes 2,888 unique images depicting 23,958 annotated objects. On average, each image contains 8.49 annotated objects, each of which has on average 8 properties. As mentioned above, the scenarios of Q-COCO consist of the bounding boxes (BBs) extracted from these images and their object/property annotations. Figure~\ref{fig:distribution} reports the distribution of scenarios with respect to the number of annotated objects included. As can be noted, scenarios containing up to 10 objects are the vast majority (around 83\% of the total).


Using these annotations, for each of the 2,888 unique scenarios, we
generate all possible queries and corresponding ground-truth answers
following the ratios defined above. To avoid including implausible
queries like, e.g., `metallic banana', when generating queries whose
answer is \textit{no}, we ensure that only properties which occur
together with the target object in at least one annotation are
included. Fig~\ref{fig:cocodata} shows some of the queries
generated from the annotation of one real image included in our
dataset.


In total, 58,673 queries are generated (mean: 20.32 queries per
image). Out of all 58,673 queries, the most represented quantifier
turns out to be \textit{no} (31,222 cases), followed by \textit{some}
(13,313), \textit{few} (9,009), \textit{most} (3,501), and
\textit{all} (1,628).  We balance their distribution when creating the
various experimental settings against which we evaluate the models
(see \S~\ref{sec:settings} for more detail).

\begin{table}[]
\centering
\begin{tabular}{lll}
\hline
                             & \textbf{Q-COCO} & \textbf{Q-ImageNet} \\ \hline
unique objects               & 29                          & 161                              \\
unique properties            & 44                          & 24                               \\
properties per object (mean) & 15.7                        & 8.0                              \\
objects per property (mean)  & 10.34                       & 53.67                            \\
objects per scenario (mean)     & 8.49                        & 16                                \\
objects per scenario (min-max)  & 6 - 22                       & 16 - 16                             \\
BBs per object (mean)        & 826.14                      & 48.38                            \\
BBs per object (min-max)     & 16 - 4741                    & 13 - 1149                         \\
BBs per property (mean)      & 2,090.39                     & 728.12                           \\
BBs per property (min-max)   & 616 - 8,320                   & 23 - 2,689                         \\
total images                 & 2,888                        & 7,790                             \\
total BBs                    & 23,958                       & 7,790                             \\ 
total queries                    & 58,673                       & 40,000                             \\ \hline
\end{tabular}
\caption{Descriptive statistics for Q-COCO and Q-ImageNet datasets.}
\label{tab:datasets}
\end{table}

\begin{figure}
\begin{center}
\hfill
\subfigure{\includegraphics[width=0.49\linewidth]{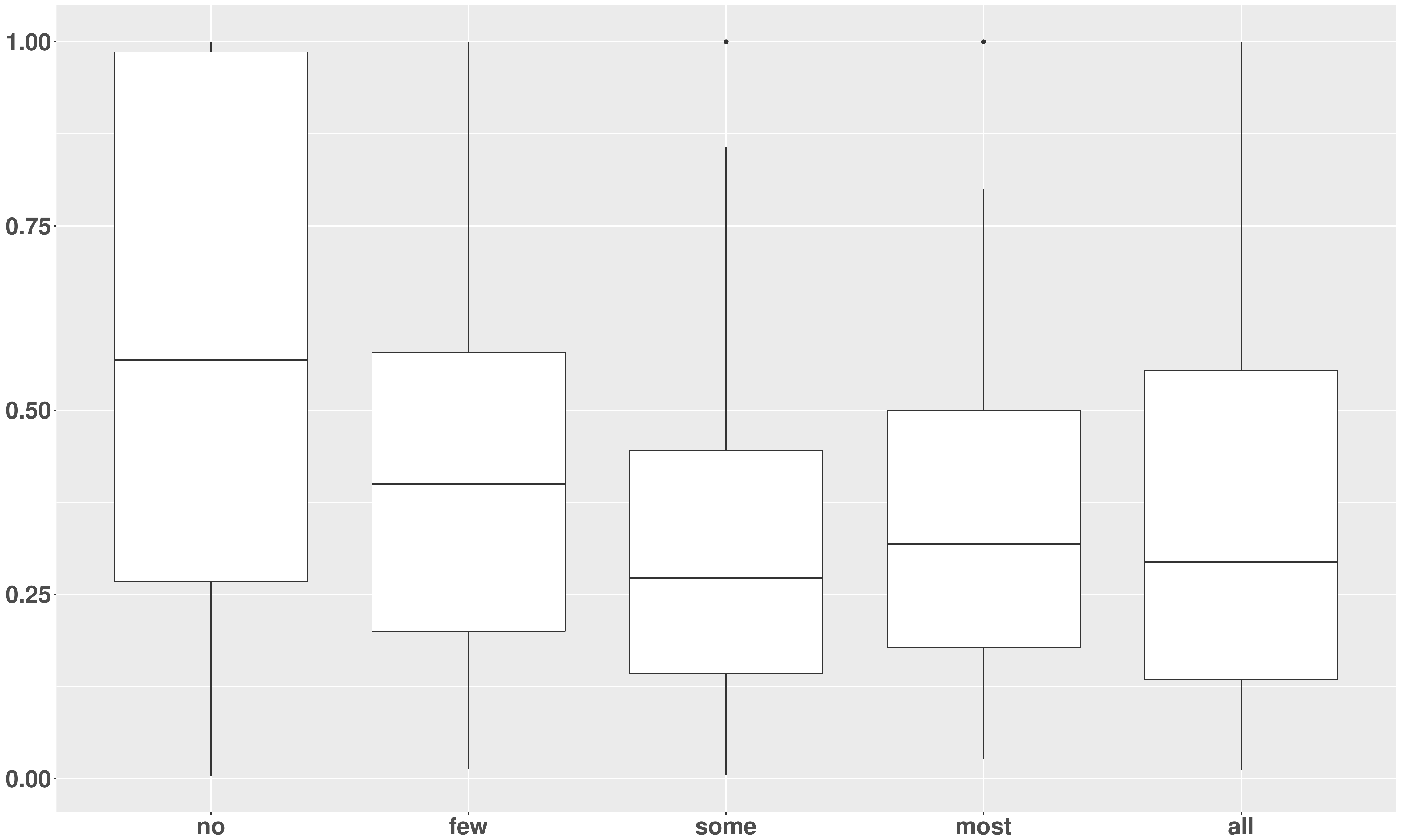}}
\hfill
\subfigure{\includegraphics[width=0.49\linewidth]{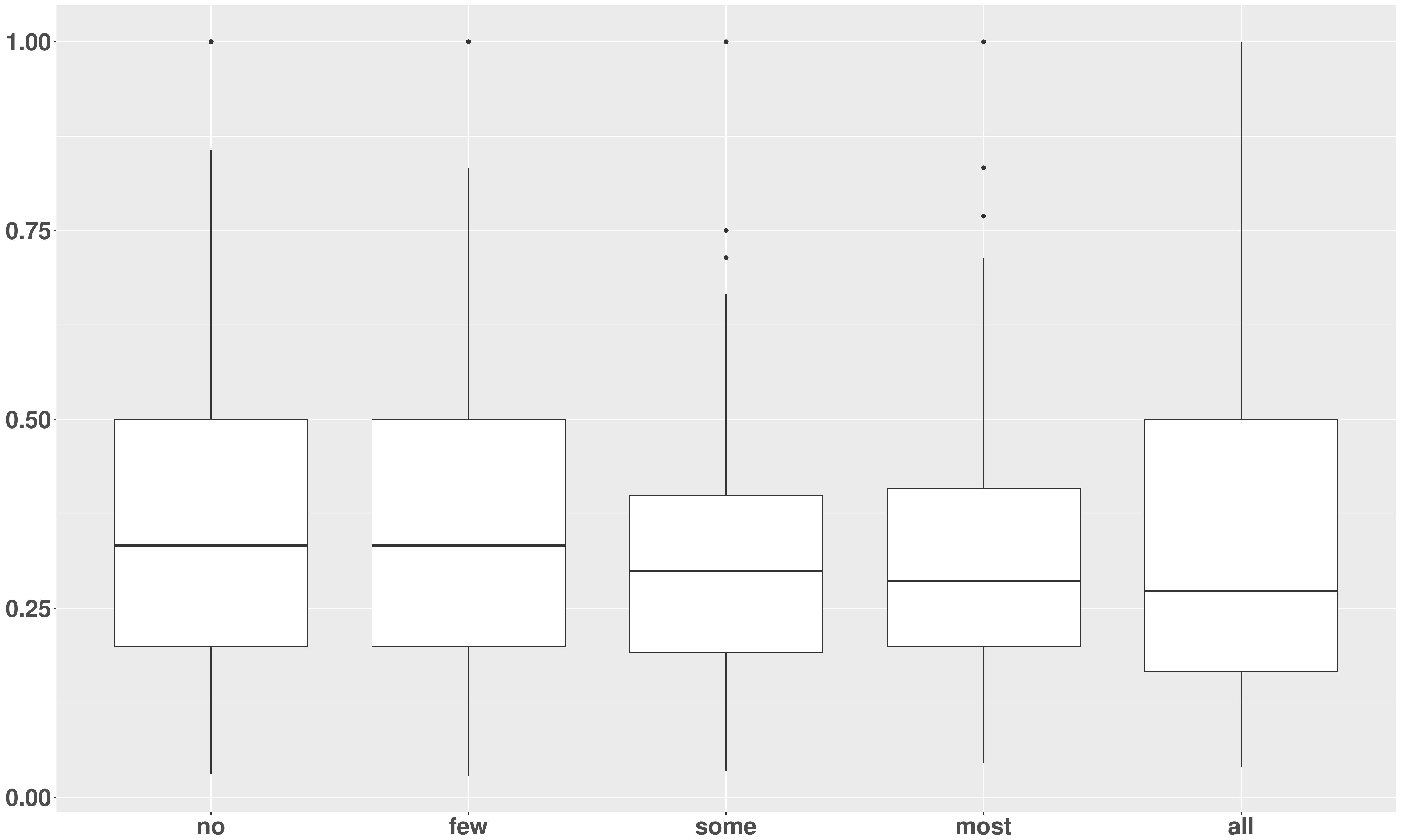}}
\hfill
\end{center}
\caption{Distributions of datapoints with respect to their frequency with a given quantifier in Q-COCO (left) and Q-ImageNet (right).}\label{fig:bias}
\end{figure}

As the literature has shown (\S\ref{sec:related_work}), datasets of natural images can be biased towards the linguistic modality.  To check whether this apply to Q-COCO, we analyse a sample of its datapoints by randomly selecting a balanced number of cases for each quantifier.  For each query (e.g., `black, dog') we compute the number of times it occurs paired with a given quantifier, e.g. `black, dog, all', in the sample dataset. We then divide this frequency by the total number of times the query `black, dog' appears in the sample dataset. This way, we obtain a \textit{ratio} describing the bias of each query toward each quantifier. If `black, dog' appears 10 times in the dataset, and these 10 cases are equally split among the 5 quantifiers (2 cases for `no', 2 cases for `few', and so on),
then the dataset can be considered as perfectly balanced, having around 20\% of cases per each quantifier. If most cases correspond to one or few specific quantifiers, then the dataset is biased. In Fig~\ref{fig:bias} (left) we plot the distribution of these ratios
relative to each quantifier.


As can be seen, \textit{no} and \textit{few} cases are particularly biased, meaning that a model could simply learn correlations between object-properties and quantifiers in order to give the right answer when tested with a seen query.  This limitation of the dataset cannot be easily solved, since any real-image dataset is likely to contain correlations that depend on object-property relations. To illustrate, `banana, metallic' -- if present -- is likely to appear with the quantifier \textit{no} (and perhaps \textit{few}), but not with \textit{most} and \textit{all}. This finding illustrates a general issue, since carrying out quantification tasks using real images might always be affected by such regularities between object-property distributions in the real world. But, as we argued in the introduction, quantifiers per se are logical functions that can in principle apply to sets of any composition.


Given the inherent bias in the object distribution of real images, we also investigate the use of a synthetic dataset. To do this, we select ImageNet as background visual corpus, since it contains more object categories and all annotated properties are visual (compare with COCO-Attribute, where properties are not necessarily visual: see Fig~\ref{fig:cocodata}).

\subsection{From ImageNet to Q-ImageNet}

ImageNet~\cite{deng2009imagenet} contains 1,073,739 images annotated with bounding boxes. Out of 3,000 object categories (`synsets' in ImageNet terminology), 375 are also provided with human annotations of properties, representing a total of 25 unique attributes. ImageNet images are rather different from those in COCO Attribute: most of the time, they don't contain multiple objects.



As in Q-COCO, we create Q-ImageNet scenarios from the bounding boxes in the image. But only one bounding box is extracted from each image. As a result, the dataset differs from
Q-COCO in that it merges together bounding boxes that do not originally belong to the same image, giving us more leeway to overcome the bias found in real scenes. 

We build synthetic scenarios that are made up by 16 different BBs. This choice is motivated by two reasons. First, in Q-COCO, 99\% of images contain 16 or less objects and so 16 can be considered as a reasonable `realistic' upper bound. Second, this number allows us to have a fairly large variability with respect to the cardinalities of both restrictor and scope.

We use the 375 objects associated with the 25 annotated property labels, and the corresponding images. We then select all ImageNet items annotated with at least one of those properties and extract the bounding box for which the human annotation has been performed. This results in 9,597 bounding boxes. This set is subsequently filtered according to the criterion that the property words must occur at least 150 times in the UkWaC corpus~\cite{baroni2009wacky}: this ensures the quality of the corresponding word embeddings. As reported in Table~\ref{tab:datasets}, after this filtering process, we end up with 161 different objects associated with 7,790 bounding boxes, and labelled with 24 properties. On average, each object has 48.38 unique bounding boxes, it is assigned 8 properties and each property is shared by 53.67 objects.

As in Q-COCO, we use our set of objects-properties to construct $\langle scenario, query, answer \rangle$ datapoints.  Since we do not start from a real image anymore, we generate a query by randomly choosing the label $l$ of one of the 161 objects and a property $p$ out of the 24 properties. In doing so, we follow the same plausibility constraint used for the previous dataset, according to which we only use $\langle object, property \rangle$ pairs that occur together at least once in the annotated images. We then assign one ground-truth answer to each scenario-query combination. Further, to make our synthetic scenarios as plausible as possible, we also set a constraint on the distractor images in each datapoint. We use an association measure based on MS-COCO captions~\cite{lin:micro14}, which evaluates the chance of two objects to appear together in a real image. The idea is that objects that are more likely to occur together make more realistic scenarios and should thus be preferred in the generation process (for instance, a dog and a sofa are more often seen together than a sofa and an elephant). We compute PMI as a proxy for the likelihood of two objects to co-occur in an image:

\begin{equation}
PMI(o1, o2) = log\frac{f(o1, o2) * N}{f(o1) * f(o2)}
\end{equation}

\noindent where $o1$ and $o2$ are two objects, $f(o1,o2)$ is the number of times \textit{words} $o1$ and $o2$ co-occur in a single caption, $f(o)$ is $o$'s frequency in the captions of MS-COCO overall, and $N$ is the number of words in all captions. If an object's label does not occur in the captions, it is considered to have the same probability of co-occurrence with all other objects.\footnote{To those unseen pairs, we assign a PMI of 0.01 -- the lowest PMI for seen pairs is 0.46 (`cheese, grass').} When selecting distractors for the object of interest in a particular scenario, we randomly pick them according to their likelihood of co-occurrence with that object, as given by the PMI calculation.

\begin{figure}
\begin{center}
  \includegraphics[width=0.9\linewidth]{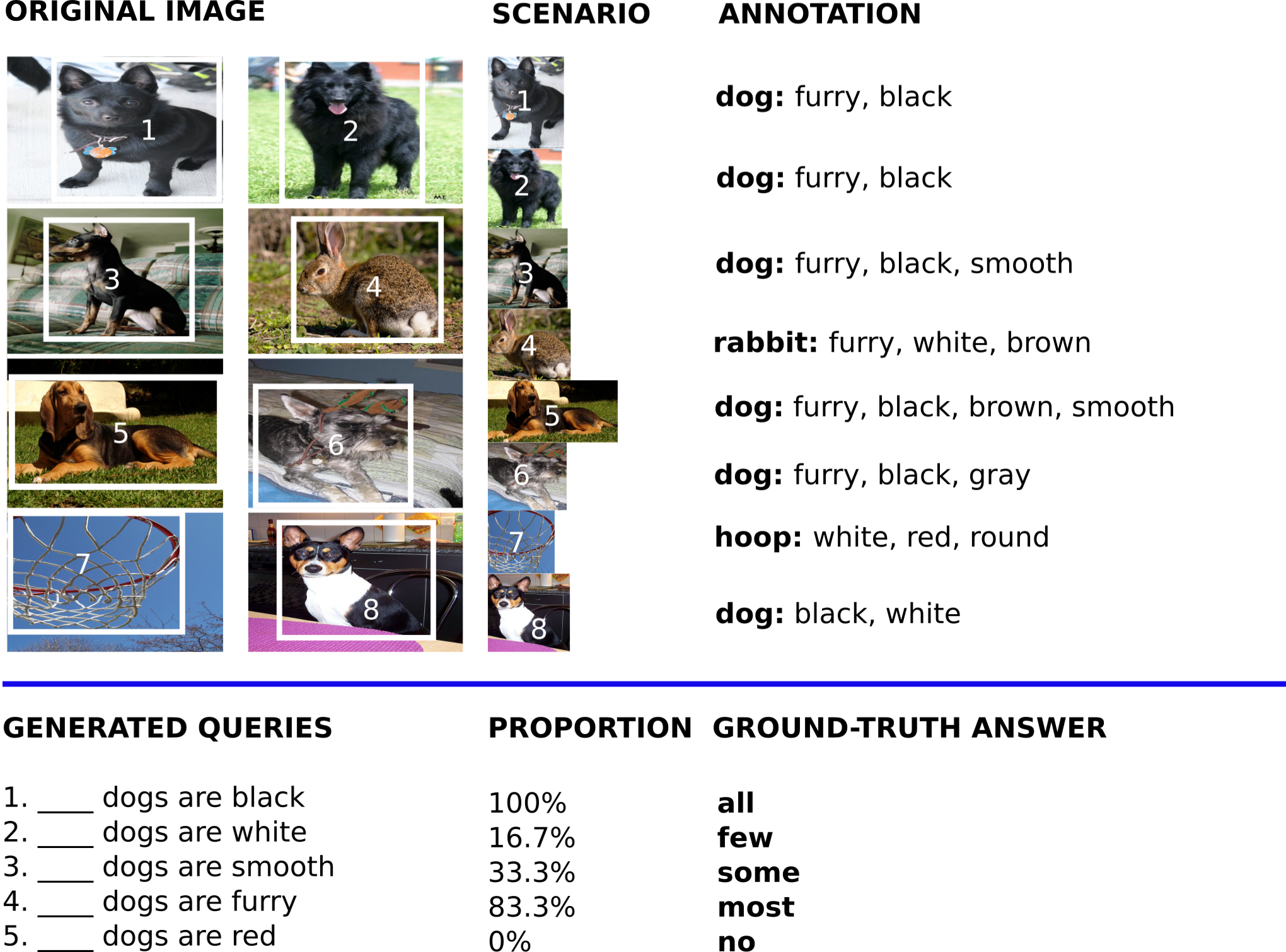}
\end{center}
\caption{Generation of datapoints in Q-ImageNet dataset. Note that, for sake of clarity, we represent the scenario as made up by 8 images instead of 16.}
\label{fig:imgdata}
\end{figure}


We check whether Q-ImageNet contains language bias by applying the same method used for Q-COCO.  In Fig~\ref{fig:bias} (right) we plot the distribution of the datapoints with respect to the proportion of cases a given query does occur with a given quantifier. As can be noticed, the distribution is much better compared to the real-image dataset. On average, our datapoints are always around chance level (i.e. 20\%), indicating that there is an almost equal number of cases for each quantifier to occur with a given query.

\section{Experimental Settings}\label{sec:settings}

For both datasets, we experiment with four experimental settings which let us test the behaviour of the system under different training conditions. 

\paragraph{Uncontrolled (UNC)} From the whole set of generated datapoints, we randomly select a balanced number of cases for each quantifier. In this setting, it is possible to encounter known scenarios or queries at test time, but scenario-query combinations are all unseen. (This setting is basically the sampled data used to control dataset bias in the previous section.)

\paragraph{Unseen objects (UnsObj)} This setting tests the generalisation power of our models over scenarios containing unseen objects. We randomly divide our list of concepts and pick 70\% for training, 30\% for testing/validation. For each concept, we then randomly select a balanced number of differently quantified datapoints. 

\paragraph{Unseen properties (UnsProp)} Similarly to UnsObj, this setting tests the generalisation power of our models with respect to properties. The procedure followed to obtain the dataset used  in this setting is the same as for the UnsObj setting, except that  we split the datapoints according to the properties. 

\paragraph{Unseen queries (UnsQue)} The last setting tests generalisation with respect to  unseen combinations $\langle object, property \rangle$. For instance, the system sees both \textit{dog} and \textit{black} in training, but $\langle dog, black \rangle$ only at test time. To build this setting, we first randomly  select 70\% $\langle object, property \rangle$ tuples for training and 30\%  tuples for testing and validation. We then follow the same procedure as above.

Details of the composition of the training, validation and test sets are given in Table~\ref{tab:settingssta}.

\begin{table}
\begin{center}

\begin{tabular}{llccccccc}\hline
& & \multicolumn{3}{c}{\textbf{Q-ImageNet}} &  & \multicolumn{3}{c}{\textbf{Q-COCO}} \\ \hline
   &                & \textbf{Train} & \textbf{Val} & \textbf{Test} & & \textbf{Train} & \textbf{Val} & \textbf{Test}\\ \hline
\multirow{2}{*}{UNC} & Datapoints &  7,000   &  1,000      & 2,000 & &
                                                                       5,600   &  800      & 1,600 \\
& Queries & 3,040 & 824 & 1,448 & & 858  & 392 & 547 \\
\multirow{2}{*}{UnsObj} & Datapoints &  7,000   &  1,000      & 2,000 &
                    & 4,050   &  450      & 900 \\
 & Object & 113 & 48 & 48 & & 19 & 10 & 10\\
\multirow{2}{*}{UnsProp} & Datapoints &  7,000   &  1,000      & 2,000 &
                    & 6,000   &  700      & 1,400 \\
& Properties & 14 & 8 & 8 & & 29 & 15 & 15\\
\multirow{2}{*}{UnsQue} & Datapoints &  7,000   &  1,000      & 2,000 &
                    & 6,100   &  600      & 1,340 \\
& Queries & 893 & 309 & 351 & & 276 & 99 & 118\\\hline
\end{tabular} 

\caption{Q-ImageNet and Q-COCO training, validation and test sets of the three experimental settings.}\label{tab:settingssta}
\end{center}
\end{table}





\section{Models}
\label{sec:models}

We experiment with seven different models, to try and understand the
contribution of various mechanisms and architectures in the
quantification task. The first two models, `blind' \textbf{BOW} and
\textbf{BOW+CNN}, are simple baselines from the VQA literature
(adapted from~\cite{zhou:simp15}). They show how a language-only model
performs over one-hot representations, and over a simple concatenation
of one-hot language vectors and CNN image representations. The next
two models, `blind' \textbf{LSTM} and \textbf{LSTM+CNN}, check on the
contribution of sequential processing to the task, both in a
language-only system and using both modalities. We expect the
sequential processing to somewhat account for the composition of the
restrictor and scope components of the query, whereas it should not
play a relevant role for the visual inputs since they are sets of
bounding boxes in which the order is not relevant.  We then turn to
attention mechanisms and adapt the Stacked Attention Network
(\textbf{SAN}) of~\cite{YangHGDS15}, hoping that attention will allow
the system to focus on relevant sets of individuals when
quantifying. Using insights from formal linguistics, we also propose a
model, the Quantification Memory Network (\textbf{QMN}), which clearly
creates separate representations for scope and restrictor of the
quantifier, following our hypothesis that quantification operates over
defined set representations. Finally, we try
to combine insights from all the investigated models in a general
system which we name Quantification Stacked Attention Network
(\textbf{QSAN}). QSAN can be seen as a linguistically-motivated
architecture based on SAN and specifically designed for quantification
task.

\subsection{Vector Representations} 
All the models receive as input `frozen' visual and linguistic
representations, obtained as follows.

\paragraph{Visual input} For each bounding box in each scenario, we
extract a visual representation using a Convolutional Neural
Network~\cite{simonyan2014very}. We use the VGG-19 model pre-trained
on the ImageNet ILSVRC data~\cite{russakovsky2015imagenet} and the
MatConvNet~\cite{matconvnet} toolbox for features extraction. Each
bounding box is represented by a 4096-dimension vector extracted from the
$7$th fully connected layer (fc7). For computational efficiency, we
subsequently reduce the vectors to 400 dimensions by applying Singular
Value Decomposition (SVD).

\paragraph{Linguistic input} Similarly, each word in a query is
represented by a 400-dimension vector built with the Word2Vec CBOW
architecture~\cite{mikolov2013}, using the parameters that were shown
to perform best in~\cite{baroni2014don}. The corpus used for building
the semantic space is a 2.8 billion tokens concatenation of the
web-based UKWaC, a mid-2009 dump of the English Wikipedia, and the
British National Corpus (BNC).

\subsection{Baselines: BOW and BOW+CNN}

As baselines, we consider two models which have shown remarkable
accuracy on the VQA task, given their simplicity: BOW and
iBOWIMG~\cite{zhou:simp15}.\footnote{Available from
  \url{https://github.com/metalbubble/VQAbaseline/}.} We implement
minor adaptations of those models to suit the quantification task, as
described below.

\paragraph{`Blind' BOW}
This is a language-only model. The network has an input layer which
has the size of the overall vocabulary (in our case, all concepts and
properties in our datasets). The query (e.g. \textit{black dog}) is
first converted to a one-hot bag-of-words (BOW) vector (activating the
units for \textit{black} and \textit{dog} in the input layer), which
is further transformed into a `word feature' embedding of 400
dimensions. The combined features are sent to a softmax layer which
predicts the answer by assigning appropriate weights to an output
layer, where each node corresponds to one of our five quantifiers.

\paragraph{CNN+BOW} 
This model is an adaptation of iBOWIMG. It uses the same linguistic
input as BOW above, concatenated with a visual input. As in BOW, the
query question is first converted to a one-hot bag-of-words vector,
which is further transformed into a `word feature' embedding. This
linguistic embedding is concatenated with an `image feature' obtained
from a convolutional neural network (CNN). The resulting embedding is
sent to a softmax classifier which predicts one of five quantifiers,
as above. In order to have one single vector for the visual input, we
simply concatenate the visual vectors of the individual bounding boxes
in each one of our scenarios. For the Q-COCO dataset, where the number
of objects contained in one images ranges from 6 to 22, we concatenate
our `frozen' visual vectors into a 8,800-dimension vector (i.e. 22*400
dimensions) and we fill the `empty' cells of the scenario with zero
vectors. For the Q-ImageNet dataset, where the number of objects is
fixed to 16, we concatenate our visual vectors into a 6,400-dimension
vector (i.e. 16*400 dimensions).

\subsection{The role of sequential processing: LSTM and LSTM+CNN }

\begin{figure}[t]
\begin{center}
  \includegraphics[width=0.9\linewidth]{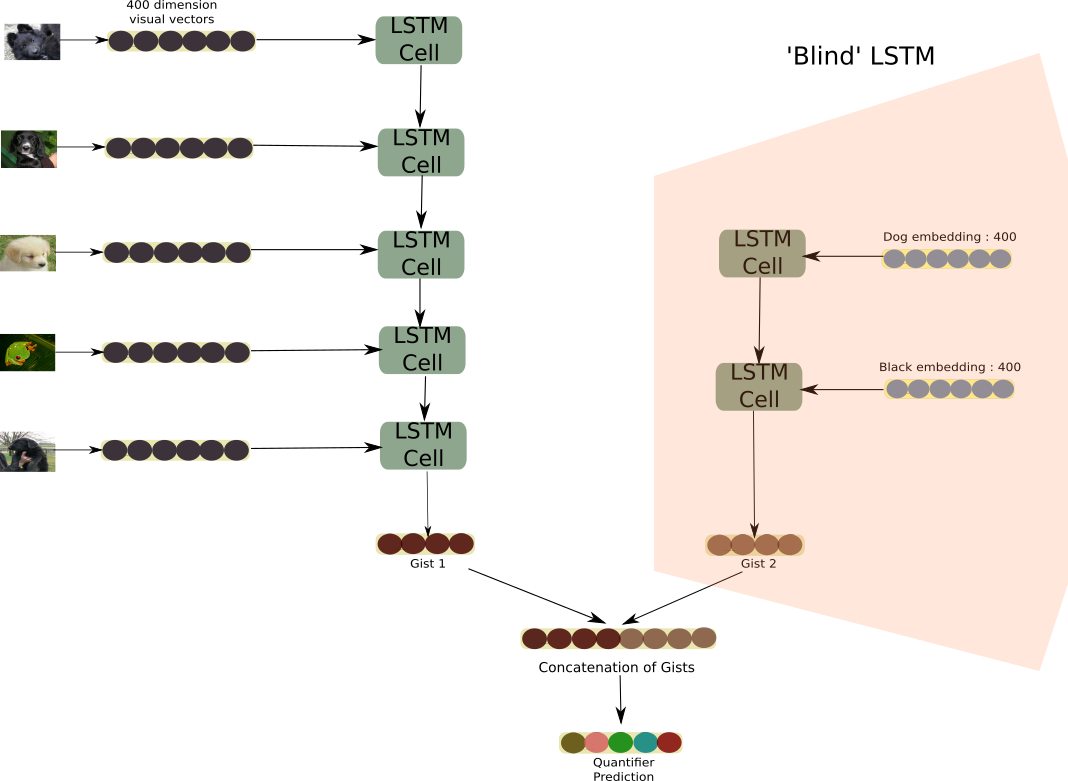}
\end{center}
\caption{Graphic representation of the CNN+LSTM model. The rightmost part included in the pink box represents the `Blind' LSTM model.}
\label{fig:cnnlstm}
\end{figure}

\paragraph{`Blind' LSTM}
A graphic representation of LSTM is provided in
Fig~\ref{fig:cnnlstm} (pink box). This model receives as input the
linguistic embeddings for each query. Then, the input is processed by
an LSTM module with two cells, which we hope might simulate the
  composition of the restrictor and scope components of the
  query; its output is linearly mapped into a 5-dimension
vector. A softmax classifier is applied on top of this vector in order
to output the correct quantifier.

\paragraph{CNN+LSTM}
As shown in Figure~\ref{fig:cnnlstm}, CNN visual features are
processed by an LSTM, with the output of the last cell (Gist1) being
combined with the linguistic information provided by the `Blind LSTM'
module processing the query (Gist2). Gist1 and Gist2 are concatenated
into a single vector on top of which a softmax classifier is applied
to output the quantifier with the highest probability.

\subsection{The role of attention mechanisms: SAN}

\paragraph{Stacked Attention Network (SAN)}


\begin{figure}[t]
\begin{center}
  \includegraphics[width=0.9\linewidth]{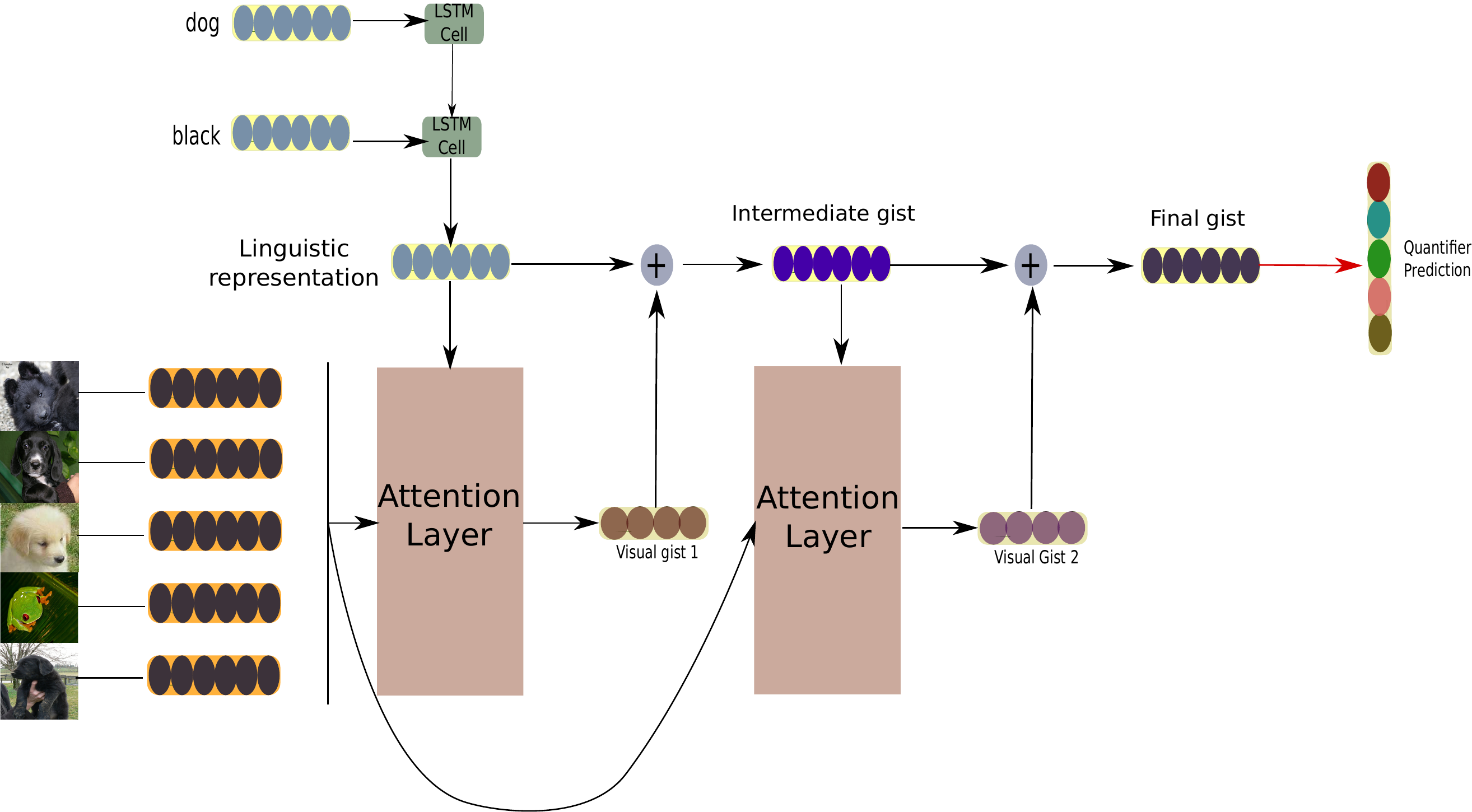}
\end{center}
\caption{Graphic representation of SAN model.}
\label{fig:san}
\end{figure}

\begin{figure}
\begin{center}
  \includegraphics[width=0.9\linewidth]{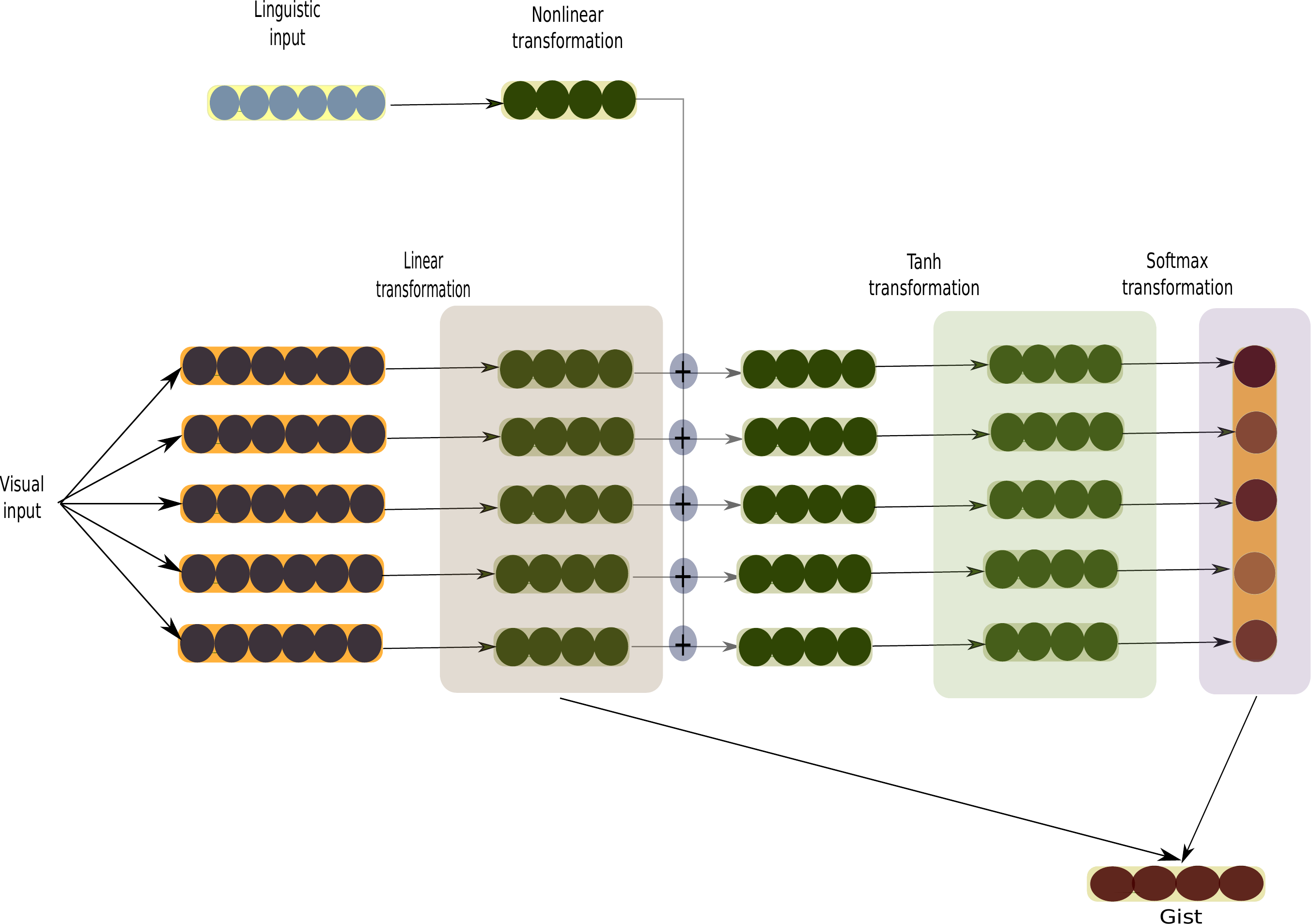}
\end{center}
\caption{Graphic representation of the SAN's attention layer.}
\label{fig:module}
\end{figure}

The Stacked Attention Network (SAN) proposed by~\cite{YangHGDS15} is
motivated by the idea that VQA might require more than one step of
reasoning. The model is supposed to pay particular attention to the
image regions that are relevant to the query via the \textbf{attention
  layer}.  The diagram presented in Figure~\ref{fig:module} zooms into
the main module of the network: the \textbf{attention layer}.  This
layer sums each visual vector with the linguistic representation and
then applies a tanh and softmax functions to the result, to obtain a
weighted average of the initial visual vectors (`gist'). The gist thus
encodes information about both the question and the
image. Consistently with the purpose of this architecture, namely
performing a multi-step reasoning, the attention layer is used twice
in SAN. As shown in Fig~\ref{fig:san}, a first pass applies the
representation of the query, as obtained from a LSTM module, to the
visual input. In the second pass, the main module takes as linguistic
input the sum of the original linguistic representation and the output
from the first pass. The final gist is then fed into a softmax
classifier to obtain the predicted quantifier.




\subsection{The role of formal linguistic structure: QMN}
\label{sec:qmn}

\begin{figure}[t]
\begin{center}
\includegraphics[scale=0.4]{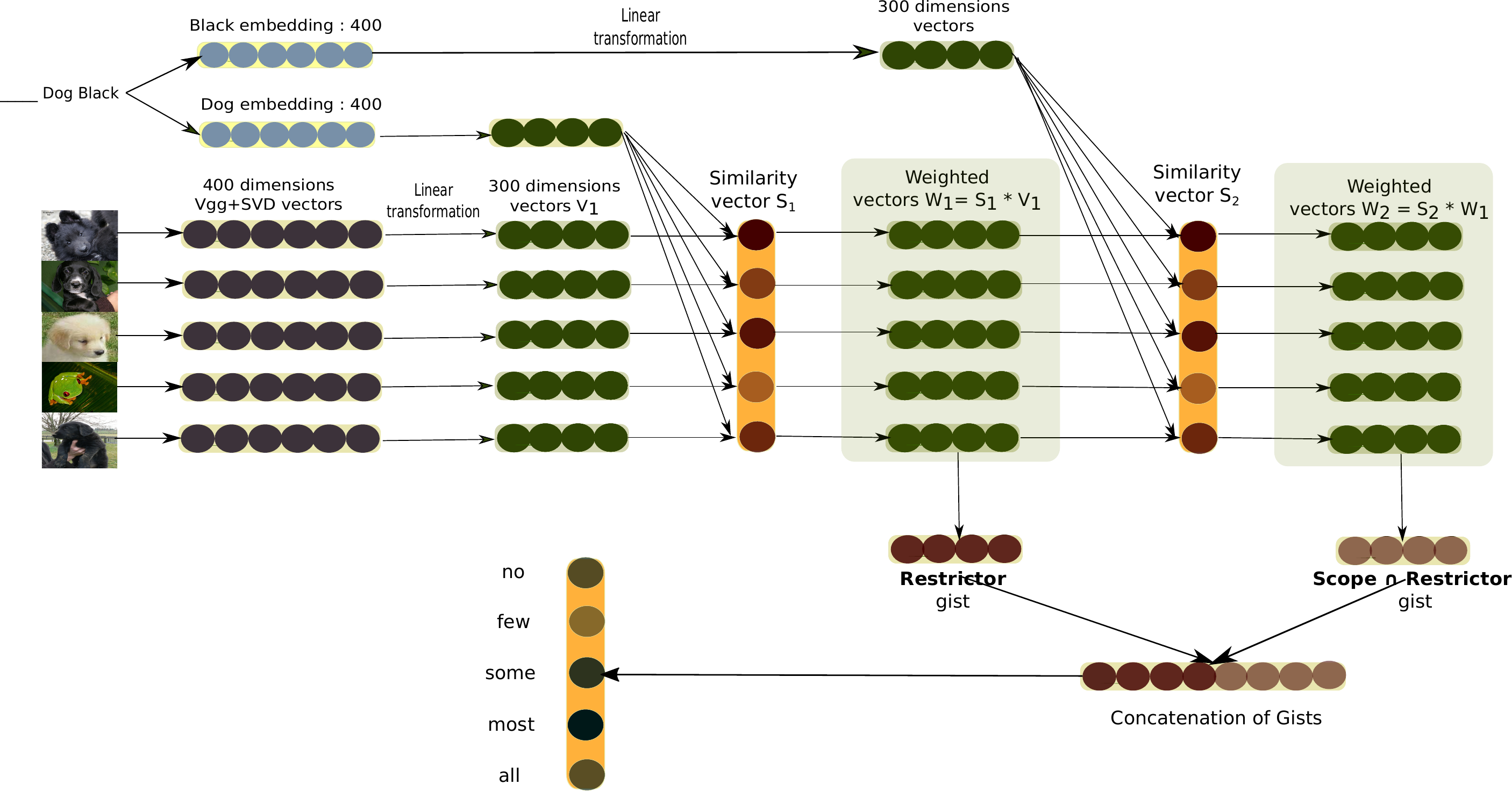}
\end{center}
\caption{A representation of the Quantification Memory Network (QMN)
  model}
\label{fig:diagram}
\end{figure}

This model is an adaptation of the Memory Network originally proposed
by~\cite{Sukhbaatar:etal:2015new}, which achieved state-of-the-art
performance in both synthetic question answering and language
modelling. The model is shown in Fig~\ref{fig:diagram}. Its main
feature is that it explicitly encodes the retrieval of the two sets
assumed by the formal semantics model of quantifiers (i.e. the
restrictor and the overlap between restrictor and scope). This model
implements our idea of a quantification model in two steps, where the
first step produces some representation of the relevant sets
of individuals, and the second step computes the relation between
those sets (see Section~\ref{sec:intro}).

\textit{Step 1:} As shown in the diagram, the visual and linguistic
vectors of all datapoints are linearly mapped to a 300-dimension
space. The 300-d visual vectors are fed into memory cells ($V1$ in
Fig~\ref{fig:diagram}); for each cell, we compute the similarity value
between each visual vector and the linguistic vector representing the
query restrictor (e.g., \textit{dog}), by calculating their dot
product further normalized using the Euclidean norm. The result is
either a 22- or 16-dimension `Similarity Vector 1' ($S1$) (for the
Q-COCO and Q-ImageNet scenarios, respectively.) in
Figure~\ref{fig:diagram}. We then calculate the weighted vectors $W1$
for each individual by multiplying the memory cells $V1$ with the
associated similarity values in $S1$. This gives us a representation
of the amount of `dogness' in each \textit{object}. The representation
of the restrictor set is calculated by summing the memory cells of the
weighted vectors obtaining the Restrictor gist. It represents how much
`dogness' is found in the given \textit{scenario}. We then calculate
the dot product between the weighted vectors ($W1$) and the scope
linguistic vector (e.g., \textit{black}), and further normalise the
values using the Euclidean norm. Again, the result is a 22- or
16-dimension `Similarity Vector 2' ($S2$). A second weighted vector
$W2$ is obtained by multiplying $W1$ and $S2$. This gives us the
amount of `black-dogness' in each \textit{object}. The representation
of the overlap between the restrictor and scope sets (Scope $\cap$
Restrictor gist) is obtained by summing the new weighted vectors in
the memory cells. It represents how much `black-dogness' is found in
the given \textit{scenario}. In this model, the
  composition of the restrictor and scope components, operationalised in the
  SAN model by the LSTM module, is accomplished by using the
  probability of the similarity vector $S2$ to weight its visual vectors $W1$.

\textit{Step 2:} The Restrictor and Scope $\cap$ Restrictor gists are
concatenated into a single 600-d vector that is further linearly
transformed into a 5-d vector. We apply a softmax classifier on top of
the resulting vector that returns the probability distribution over
the quantifiers. From the concatenation of the Restrictor gist
(`dogness') and Scope $\cap$ Restrictor gist (`black-dogness') the
model should learn the ratio between the target objects and the
restrictor and predict the quantifier that captures that relation.

\subsection{Putting it all together: QSAN}


\begin{figure}
\begin{center}
  \includegraphics[width=0.9\linewidth]{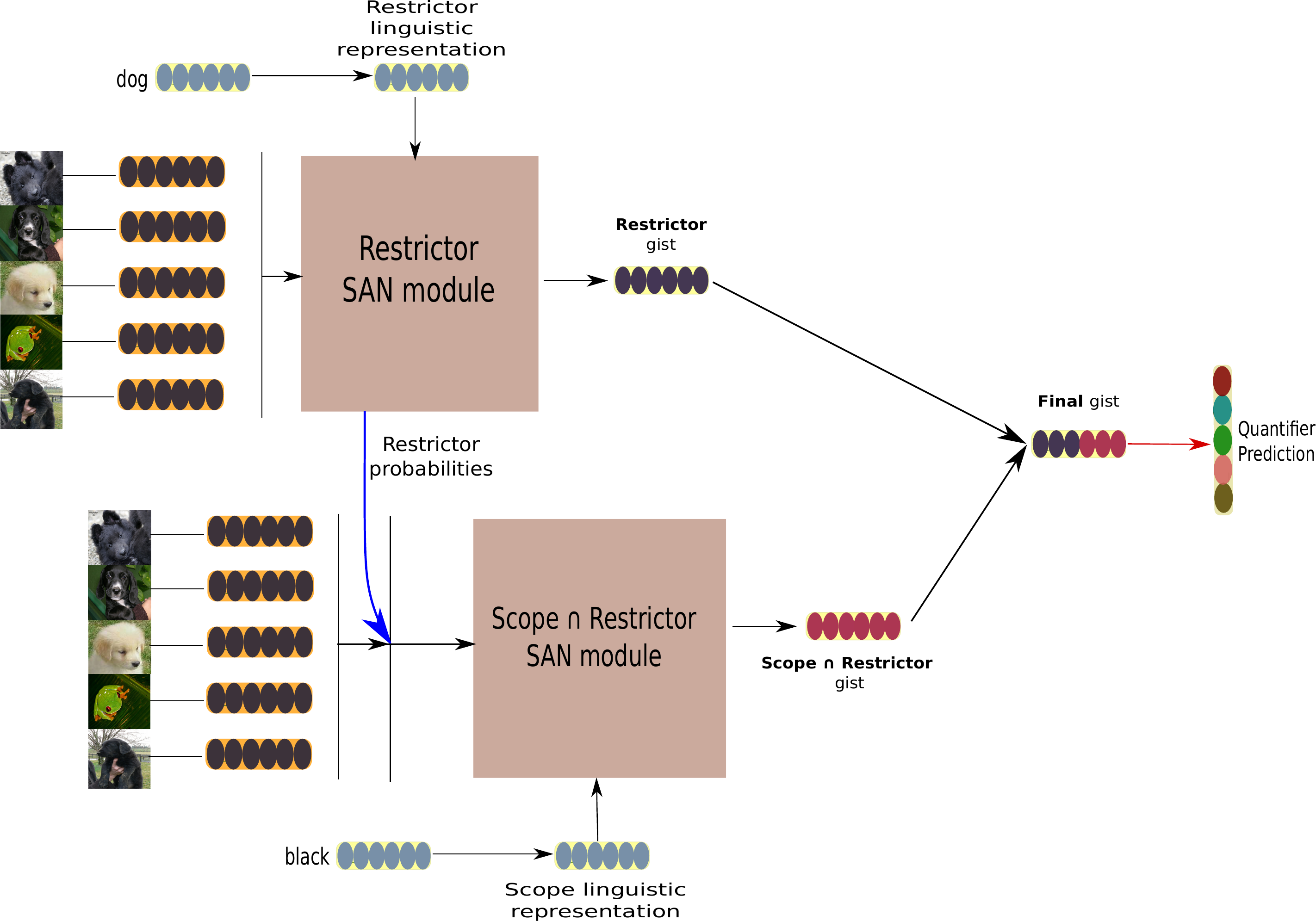}
\end{center}
\caption{Graphic representation of QSAN model.}
\label{fig:qsan}
\end{figure}


Our Quantification Stacked Attention Network (QSAN,
Fig~\ref{fig:qsan}) is an adaptation of SAN integrating the
linguistically-informed structure of the QMN. The system follows two
steps, as in the QMN.

\textit{Step 1:} the SAN model is re-implemented, with the main
difference that the given linguistic information is only the
restrictor, e.g. the embedding for the word \textit{dog}. We refer to
this part of the model as \textit{Restrictor SAN} module, and its
output as \textbf{Restrictor gist}. The network then takes the
probabilities obtained from the softmax layer in the Restrictor SAN
module, and uses these probabilities to weight the initial visual
vectors. We assume this operation will attend to the correct regions
of the visual scenario to find the restrictor set (e.g., the dogs in
the image). As in the QMN, the composition of the
  restrictor and scope is obtained by weighting the visual vectors of
  the bounding boxes with the restrictor probability before feeding them to
  the Scope Module. The weighted visual vectors are then fed into the
\textbf{Scope $\cap$ Restrictor SAN} module, where they are processed
with respect to the scope's embedding (e.g., \textit{black}). The
output of this module is the \textbf{Scope $\cap$ Restrictor gist}.

\textit{Step 2:} Restrictor and Scope $\cap$ Restrictor gists are concatenated into a
single vector on the top of which a softmax layer is applied to
predict the quantifier.

\section{Results and Analysis}
\label{sec:experiments}

In this section, we report results obtained by all models described in \S~\ref{sec:models} in all experimental settings described in \S~\ref{sec:settings}. We then zoom into more quantitative and qualitative analyses aimed at better interpreting the results.

\subsection{Results}


Results for all models in the Q-COCO settings are reported in Table~\ref{tab:coco}. The `blind' LSTM model turns out to be the best-performing model in the UNC setting (53.5\%), with the even simpler `blind' BOW achieving a remarkable good accuracy (47.3\%). This outcome is consistent with our previous discussion on the bias of this dataset towards the linguistic modality. That is, models capitalising solely on linguistic associations between objects and properties are more effective (`blind' LSTM) or similarly effective  (`blind' BOW)  as relatively complex state-of-the-art models which integrate both modalities. In other words, adding visual information does not result in any accuracy improvements in this setting. As expected, language-only models are particularly effective in predicting \textit{no} and \textit{all}, cases for which object-property distributional associations might intuitively play a higher role compared to the other quantifiers. In particular, the `blind' LSTM achieves 64\% accuracy for \textit{no} and 71\% accuracy for \textit{all}.


In UnsProp setting, all models' accuracies are around chance level. To recap, in this setting, we train the models with 29 properties and we test them with 15 unseen properties. As can be seen in Table~\ref{tab:coco}, none of the models is able to generalise to unseen properties. This confirms that the task is really challenging and it suggests that the visual information provided by CNN features tuned for the task of object classification might not be very informative as far as properties are concerned. This intuition is partially confirmed by the results for UnsObj (models are trained with 19 and 10 objects, respectively), where accuracies increase up to 30.9\% (QSAN). Even though the performance of blind models is almost the same as the best-performing QSAN model, it should be noted that generalising over unseen objects is a slightly more feasible task compared to unseen properties. Moreover, the improvement obtained by all models might be indicative of an object bias encoded in the visual vectors.

In the final setting, UnsQue (276 queries in train, 118 in test), QSAN is again the best model (42.4\%), followed by the `blind' LSTM (36.8\%) and SAN (35.4\%). The fairly large gap between the best attention network and all other models suggests that QSAN is to some extent able to generalise to unseen queries. In contrast, blind models' accuracies have a drop of more than 20\% compared to UNC, thus indicating the poor predictive power of these models when the $\langle object, property\rangle$ query has not been seen in training.





Moving to Q-ImageNet dataset, we observe in Table~\ref{tab:imagenet} that: 1) this dataset is harder than Q-COCO, since all accuracies are generally lower across all settings; 2) attention models, i.e. QSAN and SAN, turn out to be the overall best across settings, with QSAN outperforming SAN and with QMN being only slightly worse than SAN. This confirms the crucial role of using the restrictor to guide the attention through the image (and then compose) instead of composing restrictor and scope at the linguistic level only, as done by the LSTM model. In particular, QSAN model is the best-predicting in 3 settings out of 4, namely UNC, UnsObj and UnsProp, and the second-best in UnsQue. SAN is slightly worse than QSAN in UNC and UnsObj, but better than QMN. Finally, QMN outperforms both QSAN and SAN in UnsQue and it is the second-best performing in UnsProp.

\begin{figure}
\begin{center}
  \includegraphics[width=1\linewidth]{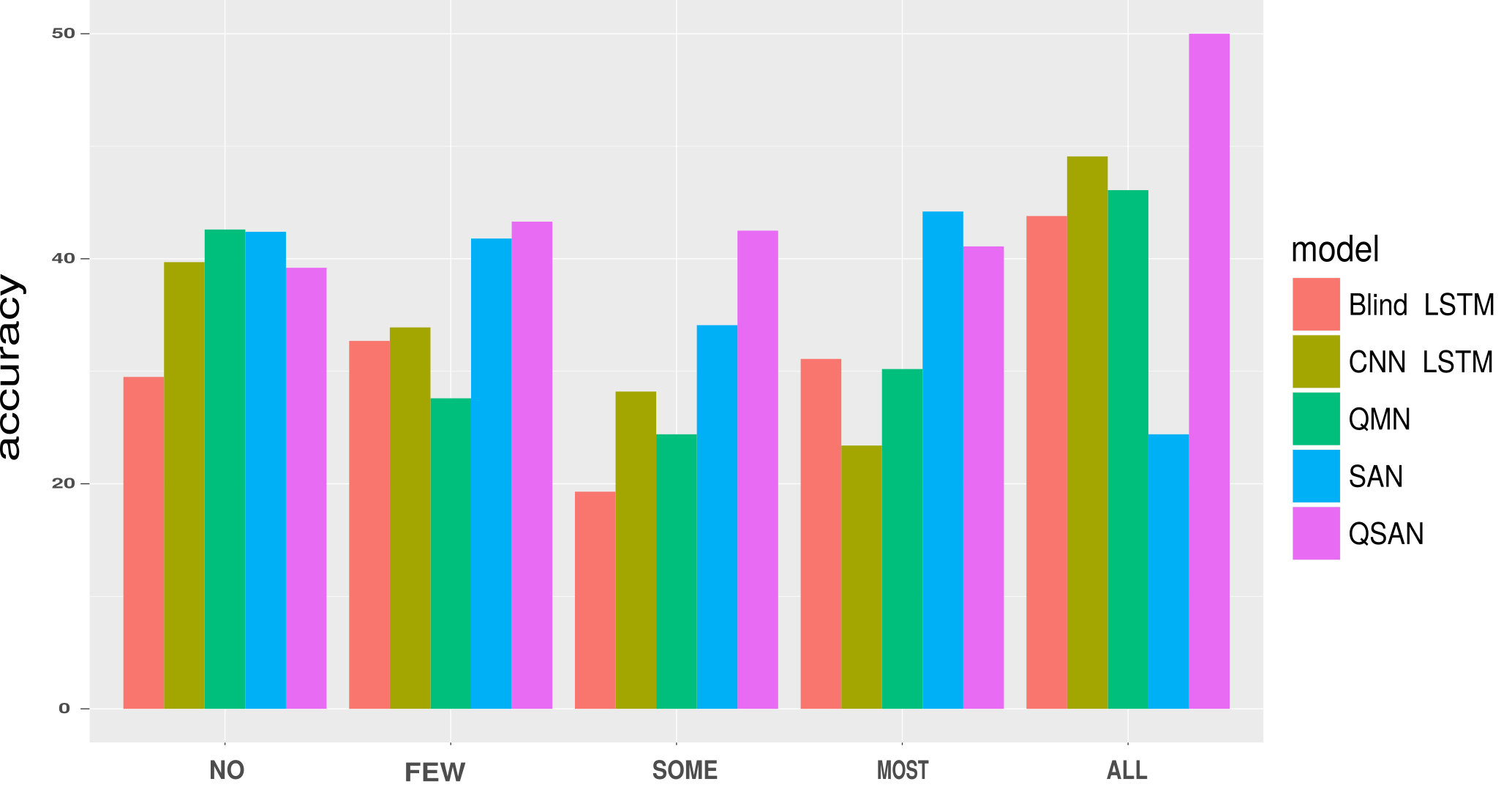}
\end{center}
\caption{UNC Q-ImageNet. Accuracies of 5 best-predicting models relative each quantifier.}
\label{fig:bigplot}
\end{figure}

Starting from UNC setting, Table~\ref{tab:imagenet} shows that QSAN outperforms SAN by almost 8\% and CNN+LSTM by almost 10\%. A visual representation of such results is provided in Fig~\ref{fig:bigplot}, which shows the accuracies of the 5 best-performing models relative to each quantifier. As can be noticed, the QSAN model outperforms the other models for \textit{few, some} and \textit{all}, whereas \textit{most} is best predicted by SAN and \textit{no} is best predicted by both QMN and SAN. At a first glance, it can be noted that on average, the accuracies of QSAN are more constant across the quantifiers, whereas all the others have some drops corresponding to specific quantifiers (see, e.g., the fairly low accuracy for \textit{all} obtained by SAN).

\begin{table}
\centering
\begin{tabular}{lcccc}
& \multicolumn{4}{c}{Q-COCO}\\\hline
& UNC & UnsObj & UnsProp & UnsQue\\\hline
Blind BOW        & 47.3                                                 & 30.7                                                    & 25.2                                                     & 34.6    \\ 
Blind LSTM & \textbf{53.5}                                        & 28.3                                                    & 25                                                       & 36.8                                                    \\ 
CNN+BOW     & 47.0                                                 & 29.8                                                    & \textbf{25.5}                                            & 33.6                                                    \\
CNN+LSTM & 49.5                                                 & 29.3                                                    & 20.8                                                     & 33.3                                                    \\ 
SAN  & 46.5                                                 & 25.9                                                    & 20.6                                                     & 35.4                                                    \\ 
QMN        & 42.7                                                 & 29.6                                                    & 23.7                                                     & 33.1                                                    \\ 
QSAN       & 51.5                                                 & \textbf{30.9}                                           & 20                                                       & \textbf{42.4}                                           \\
\textit{chance} & 20.0	& 20.0	& 20.0	& 20.0 \\ \hline
\end{tabular}
\caption{Accuracies of the various models against the Q-COCO dataset.}\label{tab:coco}
\end{table}

\begin{table}
\centering
\begin{tabular}{lcccc}
& \multicolumn{4}{c}{Q-ImageNet}\\\hline
& UNC & UnsObj & UnsProp & UnsQue\\\hline
Blind BOW        & 25.5                                                     & 25.2                                                          & 20.3	& 25.2 \\
Blind LSTM & 31.35                                                    & 23.9                                                          & 21.8                                                          & 22.3                                                          \\ 
CNN+BOW     & 26.7                                                     & 24.8                                                          & 18.9                                                          & 25.5                                                          \\
CNN+LSTM & 34.75                                                    & 23.9                                                          & 20.4                                                          & 22.8                                                          \\ 
SAN  & 37.5                                                     & 26                                                            & 20.5                                                          & 23.4                                                          \\ 
QMN        & 34.1                                                     & 23.2                                                          & 22                                                            & \textbf{28.3}                                                 \\
QSAN       & \textbf{45.2}                                            & \textbf{28.6}                                                 & \textbf{22.1}                                                 & 26 \\
\textit{chance} & 20.0	& 20.0	& 20.0	& 20.0 \\ \hline                                                           \\
\end{tabular}
\caption{Accuracies of the various models against the Q-ImageNet dataset.}\label{tab:imagenet}
\end{table}

In Table~\ref{tab:confusion} we report a quantitative analysis of the errors made in UNC by QSAN and SAN. The first thing to be noticed is that QSAN correctly predicts the target quantifier (in bold) more often than it predicts the wrong ones. In contrast, that does not hold for SAN, which predicts \textit{most} more often than \textit{all} when \textit{all} is the actual target quantifier. Second, errors made by QSAN are always `plausible', meaning that the network -- when wrong -- tends to predict quantifiers that are adjacent to the target one. That is, it wrongly outputs \textit{most} more often than \textit{some}, \textit{few}, and \textit{no} (in this order), when the target quantifier is \textit{all}. In contrast, errors in SAN do not follow the same pattern: the network indeed outputs \textit{no} more often than \textit{few} and \textit{most} when the correct quantifier is \textit{some}. Third, it should be noticed that SAN tends to be rather `negative' in its predictions, meaning that it generally outputs more answers that are `on the left' of the quantifier scale. To illustrate, it wrongly outputs more often \textit{some}, \textit{few}, and even \textit{no} than \textit{all} when the target quantifier is \textit{most}. 

As far as the other settings are concerned, a similar pattern of results as the one described for Q-COCO is observed (see Table~\ref{tab:imagenet}). In particular, all models are around chance level for UnsProp (models trained with 113 objects, tested with 48) and slightly better for UnsObj (models trained with 14 properties, tested with 8), where QSAN is the best-performing system (28.6\%), followed by the other attention model, SAN (26\%). In contrast with Q-COCO, where some models obtain fairly high accuracies in the UnsQue setting, in this dataset, none of the models reaches 30\% accuracy (models trained with 893 queries, tested with 351). QMN and QSAN are however the best (28.3\%) and second-best (26\%), respectively. The gap between the two datasets is probably due to the highest repetition of objects in Q-COCO compared to Q-ImageNet due to the comparably much lower number of object categories that are included (29 compared to 161). Even though the properties are almost halved in the latter compared to the former (24 vs 44), we conjecture that the lower number of object categories in Q-COCO plays a crucial role in helping any model to `recognise' better a given object in a scenario. Thus, having seen more often the same object in training (as in Q-COCO) should help more than having seen more often the same property (as in Q-ImageNet).


\subsection{Analysis}
\label{sec:analysis}

To better understand the results obtained with QSAN, we perform two kinds of analysis. The first is aimed at testing whether the task of predicting the correct quantifier is harder when the scenario contains an increasing number of distractors having the same queried property. For instance, if the query is \textit{black dog}, it could be the case that the model is confounded when a high number of black objects (i.e. black non-dogs) is present amongst the distractors. We check this by computing the total number of cases for each cardinality of distractors with the queried property (i.e. the number of black non-dogs) as well as the number of cases that are correctly predicted by QSAN in Q-ImageNet UNC for each cardinality. As the proportion of correctly predicted cases is constant across the various cardinalities, this factor does not seem to affect the model's performance.

\begin{figure}
\begin{center}
\hfill
\subfigure{\includegraphics[width=0.32\linewidth]{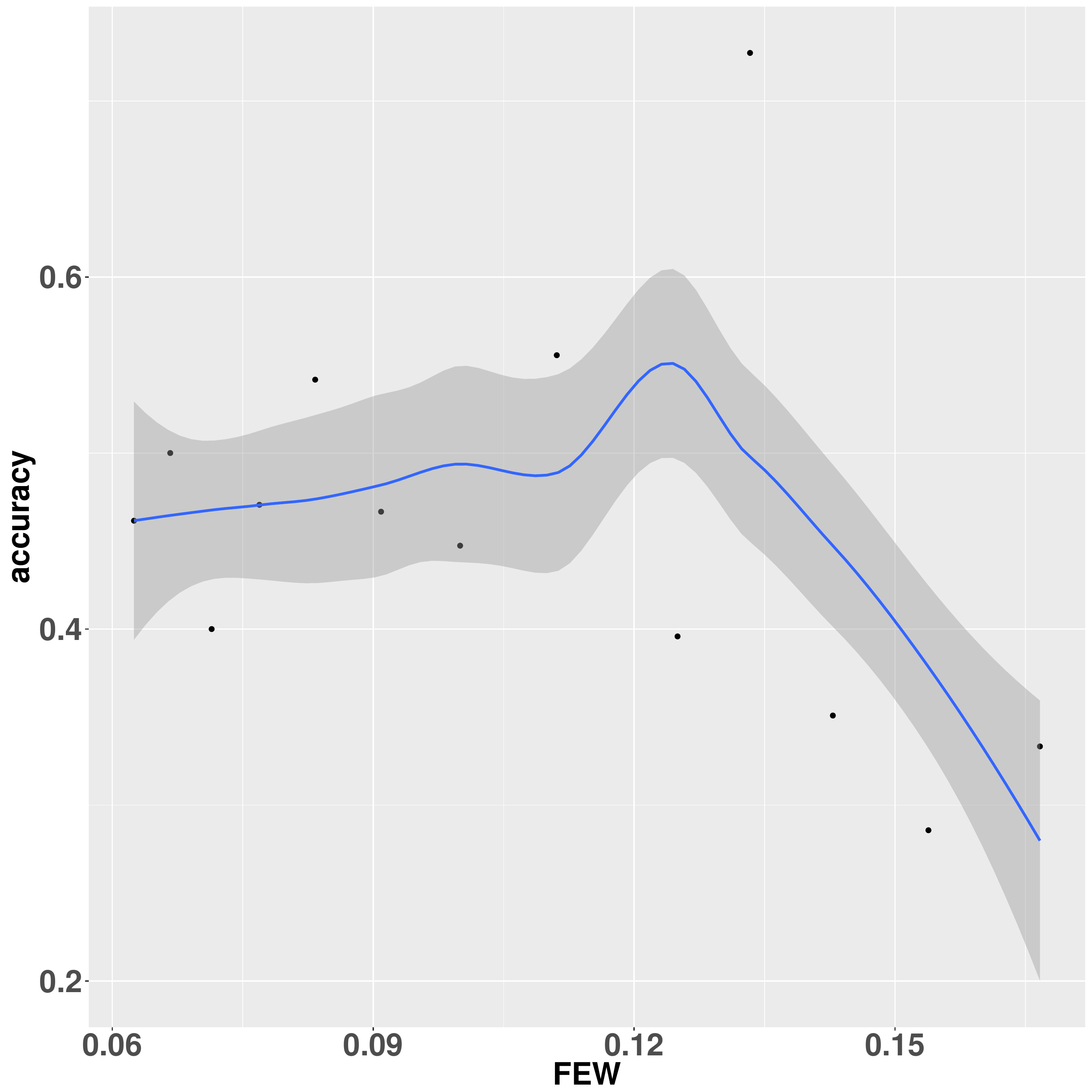}}
\hfill
\subfigure{\includegraphics[width=0.32\linewidth]{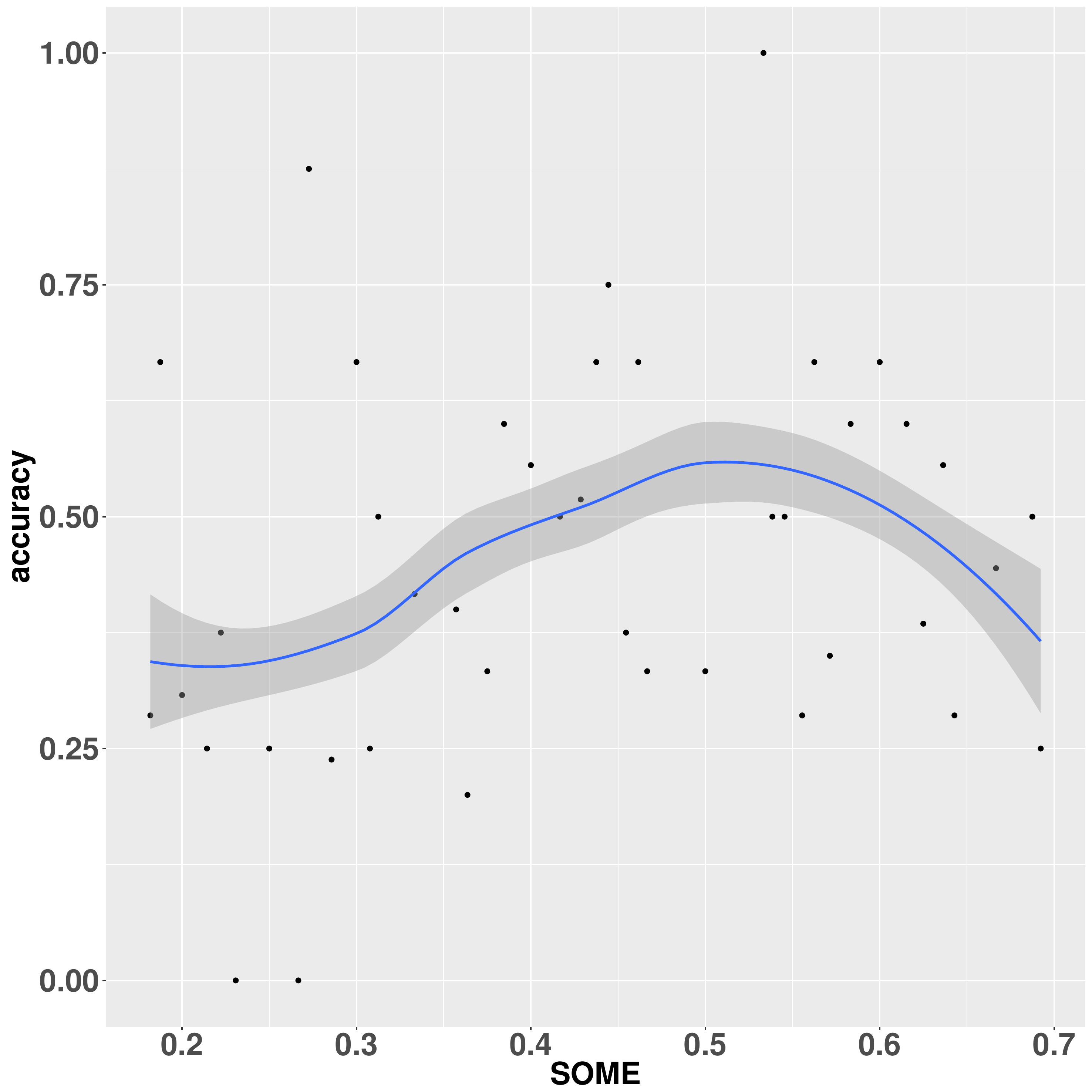}}
\hfill
\subfigure{\includegraphics[width=0.32\linewidth]{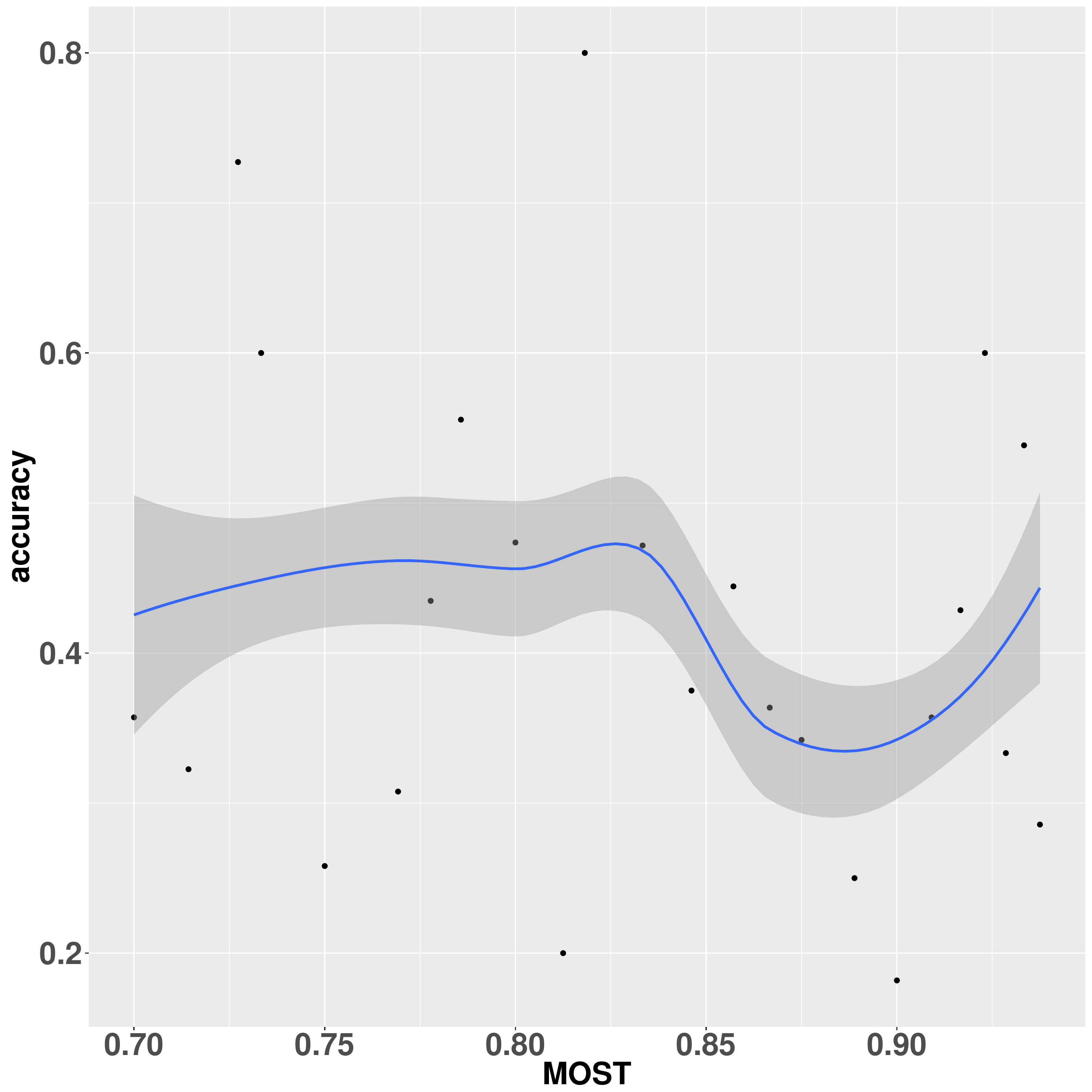}}
\hfill
\end{center}
\caption{QSAN. Accuracy in UNC plotted against the ratios of target objects over restrictors. Left: `few'. Center: `some'. Right: `most'.}\label{fig:percfsm}
\end{figure}

The second analysis is aimed at checking whether the accuracy of QSAN in Q-ImageNet UNC is affected by the actual ratio of targets over restrictor objects. Our hypothesis is that the model might be confounded with ratios that are at the boundaries between different quantifiers (e.g. across 70\%, that defines the boundary between \textit{some} and \textit{most}), while it should perform better when the ratio is undoubtedly associated with a given quantifier (e.g. around 43\% for \textit{some}). When analysing model's accuracy with respect to the whole span of ratios ranging from 0\% to 100\%, we do not find such clear `peaks'. Accordingly, model's predictions are stable across quantifiers (and relative ratios), as shown in Fig~\ref{fig:bigplot}. However, it could be the case that local patterns of fluctuation can be found within each quantifier's ratios. This is clear in Fig~\ref{fig:percfsm}, where we zoom into \textit{few} (left), \textit{some} (center), and \textit{most} (right), which are the three ones being defined by ranges. As one can notice, the expected trend is clearly visible in these plots. In particular, a peak can be observed for \textit{few} and \textit{some}, with \textit{most} having a slightly fuzzier fluctuation, that is however still consistent with our hypothesis.

\begin{table*}[]
\centering
\begin{center}

\begin{tabular}{cccccccccccc}
\hline
\multicolumn{12}{c}{\textbf{UNC Q-ImageNet}}                               \\ \hline
\multicolumn{6}{c}{\textbf{QSAN}}   & \multicolumn{6}{c}{\textbf{SAN}} \\ \hline
     & no & few & some  & most & all &       & no & few & some  & most & all \\
no & \textbf{149}  & \textit{149}  & 65  & 7  & 10   & no  & \textbf{161}  & \textit{160} & 50  & 9  & 0   \\
few  & 137   & \textbf{180} & 69  & 22  & 8   & few   & \textit{150}   & \textbf{174} & 61  & 30   & 1   \\
some   & 54   & 70   & \textbf{167} & 65  & 37   & some    & \textit{99}    & 74   & \textbf{134} & 83   & 3    \\
most  & 16   & 23  & 70 & \textbf{170} & \textit{135}   & most   & 37   & 65  & \textit{102} & \textbf{183} & 27   \\
all & 6   & 11 & 34  & \textit{108}  & \textbf{238}  & all  &  21  & 40 & 62  & \textit{177}  & 97   \\ \hline
\end{tabular}
\caption{Confusion matrices for QSAN and SAN in UNC Q-ImageNet.} \label{tab:confusion}
\end{center}
\end{table*}


\cut{\begin{figure}
\begin{center}
  \includegraphics[width=0.8\linewidth]{images/quantifiers_perc.pdf}
\end{center}
\caption{QSAN. Accuracy in UNC plotted against the full span of ratios of target objects over restrictors.}
\label{fig:perc}
\end{figure}
}

A third, more general analysis, aims at understanding to which extent quantification is made harder by having to deal with `real' concepts and images. What we wish to check is whether the purely logical part of the quantifier, which computes a ratio between two sets, can easily be learnt by a network. To do this, we reduce the uncontrolled Q-ImageNet dataset to its simplest instance, as white dots (corresponding to the intersection between restrictor and scope) and black dots (corresponding to the restrictor), in images with a gray background.

We then build a simple classifier over this data, by training from
scratch a shallow convolutional neural network (CNN), with just one
convolution layer. 
This system obtains 96\% accuracy, confirming NNs can learn the quantification comparison step nearly perfectly if a completely abstract representation is given. This is an interesting result which confirms that the actual challenge of visual quantification is to find the right strategies to deal with uncertainty in object and property recognition. As the psycholinguistic literature shows, humans appeal extensively to their approximate number sense to quantify (see \S\ref{sec:related_work}). This may be more than an efficiency mechanism: as demonstrated by the QSAN model's combination of soft attention and gist, approximation goes a long way in manoeuvring through the difficulties of matching words and vision.

\section{Conclusion}
\label{sec:conclusions}

In this paper, we investigated the task of quantifying over visual scenes using natural language quantifiers. As discussed in Section~\ref{sec:intro}, assigning a quantifier to a scenario involves two steps a) an approximate number estimation mechanism, acting over the relevant sets in the image; b) a quantification comparison step. The most straightforward and logical strategy to learn such two-step operation would be to divide the task into two subtasks: learning a correlation (a) from raw data to abstract set representation and (b) from the latter to quantifiers. The high results obtained in~\cite{soro:look16}, who have trained NNs to quantify over synthetic scenarios of coloured dots, suggest that NNs should be able to learn the second subtask quite easily. Our own experiments using a shallow CNN with just one convolution layer over abstract images confirms this. However, we know from previous work that object identification and in particular \textit{property identification} is not a solved problem. For our task, a single mistake in identification can have dramatic consequences, especially when considering sets of small cardinalities (a ratio of $\frac{1}{6}$ in our setup corresponds to \textit{few}, while $\frac{2}{6}$ is \textit{some}). It is also unclear that exact object identification is performed by humans when they quantify (see \S\ref{sec:related_work}). We therefore explored a model that is able to deal with uncertainties in both identification and cardinality estimation, and relies on soft attention mechanisms.

We first showed that letting the network compose scope and restrictor on the language side, and using this representation to attend to the image, resulted in underperforming models. Instead, using the linguistic representation of the quantifier as a relation between sets, guiding the attention mechanism, produced much better accuracy, as illustrated by the QMN and QSAN models. We take this result to show that, when considering complex, high-level phenomena, it is useful to correlate insights from formal linguistics with targeted NN mechanisms. We hope that our study will encourage further work in building linguistically-motivated neural architectures.

\bibliography{../raffa,../aurelie,../marco,../sandro}

\begin{thebibliography}{}

\bibitem[Anderson et~al., 2013]{Anderson2013}
Anderson, A.~J., Bruni, E., Bordignon, U., Poesio, M., and Baroni, M. (2013).
\newblock Of words, eyes and brains: Correlating image-based distributional
  semantic models with neural representations of concepts.
\newblock In {\em EMNLP}, pages 1960--1970.

\bibitem[Andreas et~al., 2016a]{andr:learn16}
Andreas, J., Rohrbach, M., Darrell, T., and Klein, D. (2016a).
\newblock Learning to compose neural networks for question answering.
\newblock In {\em Proceedings of NAACL-HLT 2016}, page 1545–1554, San Diego,
  California. Association for Computational Linguistics.

\bibitem[Andreas et~al., 2016b]{andre:neur16}
Andreas, J., Rohrbach, M., Darrell, T., and Klein, D. (2016b).
\newblock Neural module networks.
\newblock In {\em Proceedings of the Conference on Computer Vision and Pattern
  Recognition}.

\bibitem[Antol et~al., 2015]{anto:vqa15}
Antol, S., Agrawal, A., Lu, J., Mitchell, M., Batra, D., Zitnick, C.~L., and
  Parikh, D. (2015).
\newblock {VQA}: Visual question answering.
\newblock In {\em International Conference on Computer Vision (ICCV)}.

\bibitem[Baroni et~al., 2012]{Baroni2012}
Baroni, M., Bernardi, R., Do, N.-Q., and Shan, C.-c. (2012).
\newblock Entailment above the word level in distributional semantics.
\newblock In {\em Proceedings of the 13th Conference of the European Chapter of
  the Association for Computational Linguistics}, pages 23--32. Association for
  Computational Linguistics.

\bibitem[Baroni et~al., 2009]{baroni2009wacky}
Baroni, M., Bernardini, S., Ferraresi, A., and Zanchetta, E. (2009).
\newblock The wacky wide web: a collection of very large linguistically
  processed web-crawled corpora.
\newblock {\em Language resources and evaluation}, 43(3):209--226.

\bibitem[Baroni et~al., 2014]{baroni2014don}
Baroni, M., Dinu, G., and Kruszewski, G. (2014).
\newblock Don't count, predict! a systematic comparison of context-counting vs.
  context-predicting semantic vectors.
\newblock In {\em ACL (1)}, pages 238--247.

\bibitem[Boleda and Herbelot, 2016]{bole:form16}
Boleda, G. and Herbelot, A. (2016).
\newblock Formal distributional semantics: Introduction to the special issue.
\newblock {\em Computational Linguistics}, 42(4):619--635.

\bibitem[Borji et~al., 2014]{li:sali14}
Borji, A., Cheng, M., Jiang, H., and Li, J. (2014).
\newblock Salient object detection: A benchmark.
\newblock {\em IEEE Transactions on Image Processing}.

\bibitem[Chattopadhyay et~al., 2016]{pari:count16}
Chattopadhyay, P., Vedantam, R., Selvaraju, R.~R., Batra, D., and Parikh, D.
  (2016).
\newblock Counting everyday objects in everyday scenes.
\newblock arXiv:1604.03505.

\bibitem[Dehaene and Changeux, 1993]{deha:deve93}
Dehaene, S. and Changeux, J. (1993).
\newblock Development of elementary numerical abilities: A neuronal model.
\newblock {\em Journal of Cognitive Neuroscience}, 5.

\bibitem[Deng et~al., 2009]{deng2009imagenet}
Deng, J., Dong, W., Socher, R., Li, L.-J., Li, K., and Fei-Fei, L. (2009).
\newblock Imagenet: A large-scale hierarchical image database.
\newblock In {\em Computer Vision and Pattern Recognition, 2009. CVPR 2009.
  IEEE Conference on}, pages 248--255. IEEE.

\bibitem[Fukui et~al., 2016]{fuki:mult16}
Fukui, A., Park, D.~H., Yang, D., Rohrbach, A., Darrell, T., and Rohrbach, M.
  (2016).
\newblock Multimodal compact bilinear pooling for visual question answering and
  visual grounding.
\newblock In {\em Conference on Empirical Methods in Natural Language
  Processing (EMNLP)}.

\bibitem[Gao et~al., 2015]{gao:arey15}
Gao, H., Mao, J., Zhou, J., Huang, Z., and Yuille, A. (2015).
\newblock Are you talking to a machine? dataset and methods for multilingual
  image question answering.
\newblock In {\em International Conference on Learning Representations}.

\bibitem[Geman et~al., 2015]{gema:visu15}
Geman, D., GErman, S., Hallonquist, N., and Younes, L. (2015).
\newblock Visual turing test for computer vision systems.
\newblock {\em PNAS}, 112(12):3618--3623.

\bibitem[{Goyal} et~al., 2016]{goya:maki16}
{Goyal}, Y., {Khot}, T., {Summers-Stay}, D., {Batra}, D., and {Parikh}, D.
  (2016).
\newblock {Making the V in VQA Matter: Elevating the Role of Image
  Understanding in Visual Question Answering}.
\newblock {\em ArXiv e-prints}.

\bibitem[Herbelot and Vecchi, 2015]{Herbelot2015}
Herbelot, A. and Vecchi, E.~M. (2015).
\newblock {Building a shared world: Mapping distributional to model-theoretic
  semantic spaces}.
\newblock In {\em Proceedings of the 2015 Conference on Empirical Methods in
  Natural Language Processing}, Lisbon, Portugal.

\bibitem[Hodosh et~al., 2013]{Hodosh:etal:2013}
Hodosh, M., Young, P., and Hockenmaier, J. (2013).
\newblock Framing image description as a ranking task: Data, models and
  evaluation metrics.
\newblock {\em Journal of Artificial Intelligence Research}, 47:853--899.

\bibitem[Johnson et~al., 2017]{girs:clev16}
Johnson, J., Hariharan, B., van~der Maaten, L., Fei-Fei, L., Zitnick, C.~L.,
  and Girshick, R. (2017).
\newblock Clevr: A diagnostic dataset for compositional language and elementary
  visual reasoning.
\newblock In {\em Proceedings of CVPR 2017}.

\bibitem[Keenan and Paperno, 2012]{hand:keen}
Keenan, E. and Paperno, D., editors (2012).
\newblock {\em Handbook of Quantifiers in Natural Language}.
\newblock Springer.

\bibitem[Khemlani et~al., 2009]{khemlani2009generics}
Khemlani, S., Leslie, S.-J., and Glucksberg, S. (2009).
\newblock Generics, prevalence, and default inferences.
\newblock In {\em Proceedings of the 31st annual conference of the Cognitive
  Science Society}, pages 443--448. Cognitive Science Society Austin, TX.

\bibitem[Kumar et~al., 2016]{kuma:askm16}
Kumar, A., Irsoy, O., Su, J., J.~Bradbury, R.~E., Pierce, B., Ondruska, P.,
  Gulrajani, I., and Socher, R. (2016).
\newblock Ask me anything: Dynamic memory networks for natural language
  processing.
\newblock In {\em Proceedings of the International Conference on Machine
  Learning (ICML)}.

\bibitem[Lazaridou et~al., 2015]{Lazaridou2015}
Lazaridou, A., Pham, N.~T., and Baroni, M. (2015).
\newblock Combining language and vision with a multimodal skip-gram model.
\newblock In {\em Proceedings of NAACL}.

\bibitem[Lin et~al., 2014a]{lin:micr14}
Lin, T.-Y., Maire, M., Belongie, S., Hays, J., Perona, P., Ramanan, D., Dollar,
  P., and Zitnick, C.~L. (2014a).
\newblock Microsoft {COCO}: Common objects in context.
\newblock In {\em Proceedings of ECCV (European Conference on Computer
  Vision)}.

\bibitem[Lin et~al., 2014b]{lin:micro14}
Lin, T.-Y., Maire, M., Belongie, S., Hays, J., Perona, P., Ramanan, D., Dollar,
  P., and Zitnick, C.~L. (2014b).
\newblock Microsoft coco: Common objects in context.
\newblock In {\em Microsoft COCO: Common Objects in Context}.

\bibitem[Ma et~al., 2016]{ma:lear16}
Ma, L., Lu, Z., and Li, H. (2016).
\newblock Learning to answer questions from image using convolutional neural
  network.
\newblock In {\em Proceedings of the Thirtieth AAAI Conference on Artificial
  Intelligence(AAAI)}.

\bibitem[Malinowski and Fritz, 2014]{mali:amul14}
Malinowski, M. and Fritz, M. (2014).
\newblock A multi-world approach to question answering about real-world scenes
  based on uncertain input.
\newblock In {\em Advances in Neural Information Processing Systems}.

\bibitem[Malinowski et~al., 2015]{mali:asky15}
Malinowski, M., Rohrbach, M., and Fritz, M. (2015).
\newblock Ask your neurons: A neural-based approach to answering questions
  about images.
\newblock In {\em In International Conference on Computer Vision (ICCV'15)}.

\bibitem[Mikolov et~al., 2013]{mikolov2013}
Mikolov, T., Chen, K., Corrado, G., and Dean, J. (2013).
\newblock Efficient estimation of word representations in vector space.
\newblock {\em arXiv preprint arXiv:1301.3781}.

\bibitem[Patterson and Hays, 2016]{patterson2016coco}
Patterson, G. and Hays, J. (2016).
\newblock Coco attributes: Attributes for people, animals, and objects.
\newblock {\em European Conference on Computer Vision}.

\bibitem[Pezzelle et~al., 2017]{pezz:bepr16}
Pezzelle, S., Marelli, M., and Bernardi, R. (2017).
\newblock Be precise or fuzzy: Learning the meaning of cardinals and
  quantifiers from vision.
\newblock In {\em In Proceedings of EACL}.

\bibitem[Piantadosi, 2011]{pian:lear11}
Piantadosi, S.~T. (2011).
\newblock {\em Learning and the language of thought}.
\newblock PhD thesis, Massachusetts Institute of Technologu.

\bibitem[Piantadosi et~al., 2012]{pian:mode12}
Piantadosi, S.~T., Tenenbaum, J.~B., and Goodman, N.~D. (2012).
\newblock Modeling the acquistiion of quantifier semantics: a case study in
  function word learnability.

\bibitem[Rajapakse et~al., 2005]{raja:groun05}
Rajapakse, R., Cangelosi, A., Conventry, K., Newstead, S., and Bacon, A.
  (2005).
\newblock Grounding linguistic quantifiers in perception: Experiments on
  numerosity judgments.
\newblock In {\em Proceeding of the 2nd Language and Technology Conference},
  Poland.

\bibitem[Ren et~al., 2015a]{ren:expl15}
Ren, M., Kiros, R., and Zemel, R. (2015a).
\newblock Exploring models and data for image question answering.
\newblock In {\em Advances in Neural Information Processing Systems (NIPS
  2015)}.

\bibitem[Ren et~al., 2015b]{ren:imag15}
Ren, M., Kiros, R., and Zemel, R. (2015b).
\newblock Image question answering: A visual semantic embedding model and a new
  dataset.
\newblock In {\em In International Conference on Machine Learning Deep Learning
  Workshop}.

\bibitem[Russakovsky et~al., 2015]{russakovsky2015imagenet}
Russakovsky, O., Deng, J., Su, H., Krause, J., Satheesh, S., Ma, S., Huang, Z.,
  Karpathy, A., Khosla, A., Bernstein, M., et~al. (2015).
\newblock Imagenet large scale visual recognition challenge.
\newblock {\em International Journal of Computer Vision}, 115(3):211--252.

\bibitem[Segu{\'\i} et~al., 2015]{Segui2015}
Segu{\'\i}, S., Pujol, O., and Vitria, J. (2015).
\newblock Learning to count with deep object features.
\newblock In {\em Proceedings of the IEEE Conference on Computer Vision and
  Pattern Recognition Workshops}, pages 90--96.

\bibitem[Simonyan and Zisserman, 2014]{simonyan2014very}
Simonyan, K. and Zisserman, A. (2014).
\newblock Very deep convolutional networks for large-scale image recognition.
\newblock {\em CoRR}, abs/1409.1556.

\bibitem[Sorodoc et~al., 2016]{soro:look16}
Sorodoc, I., Lazaridou, A., ´, G. B. A.~H., Pezzelle, S., and Bernardi, R.
  (2016).
\newblock “look, some green circles!”: Learning to quantify from image.
\newblock In {\em Proceedings of the 5th Workshop on Vision and Language}, page
  75–79, Berlin, Germany. Association for Computational Linguistics.

\bibitem[Stoianov and Zorzi, 2012]{Stoianov2012}
Stoianov, I. and Zorzi, M. (2012).
\newblock Emergence of a'visual number sense'in hierarchical generative models.
\newblock {\em Nature neuroscience}, 15(2):194--196.

\bibitem[Sukhbaatar et~al., 2015]{Sukhbaatar:etal:2015new}
Sukhbaatar, S., Szlam, A., Weston, J., and Fergus, R. (2015).
\newblock End-to-end memory networks.
\newblock In {\em Proceedings of Advances in Neural Information Processing
  Systems (NIPS 2015)}, volume~28.

\bibitem[Szabolsci, 2010]{szab:quant10}
Szabolsci, A. (2010).
\newblock {\em Quantification}.
\newblock Cambridge University Press.

\bibitem[van Benthem, 1986]{vanb:essa86}
van Benthem, J. (1986).
\newblock {\em Essays in Logical Semantics}.
\newblock Reidel Publishing Co, Dordrecht, The Netherlands.

\bibitem[Vedaldi and Lenc, 2015]{matconvnet}
Vedaldi, A. and Lenc, K. (2015).
\newblock {\em MatConvNet -- Convolutional Neural Networks for MATLAB}.
\newblock Proceeding of the {ACM} Int. Conf. on Multimedia.

\bibitem[Weston et~al., 2015]{west:memo15}
Weston, J., Chopra, S., and Bordes, A. (2015).
\newblock Memory networks.
\newblock In {\em International Conference on Learning Representations (ICLR)}.

\bibitem[Xiong et~al., 2016]{xion:dyna16}
Xiong, C., Merity, S., and Socher, R. (2016).
\newblock Dynamic memory networks for visual and textual question answering.
\newblock In {\em In Proceedings of International Conference on Machine
  Learning (ICML)}.

\bibitem[Xu et~al., 2015]{xu:show15}
Xu, K., Ba, J.~L., Kiros, R., Cho, K., Courville, A., Salakhutdinov, R., Zemel,
  R., and Bengio, Y. (2015).
\newblock Show, attend and tell: Neural image caption generation with visual
  attention.
\newblock In {\em Proceedings of the International Conference on Machine
  Learning (ICML)}.

\bibitem[Yang et~al., 2016]{yang:stac16}
Yang, Z., He, X., Gao, J., Deng, L., and Smola, A. (2016).
\newblock Stacked attention networks for imagequestion answering.
\newblock In {\em In Proceedings of CVPR}.

\bibitem[Yang et~al., 2015]{YangHGDS15}
Yang, Z., He, X., Gao, J., Deng, L., and Smola, A.~J. (2015).
\newblock Stacked attention networks for image question answering.
\newblock {\em CoRR}, abs/1511.02274.

\bibitem[Zhang et~al., 2015]{zhan:sali15}
Zhang, J., Ma, S., Sameki, M., Sclaroff, S., Betke, M., Lin, Z., Shen, X.,
  Price, B., and ech, R.~M. (2015).
\newblock Salient object subitizing.
\newblock In {\em Proceedings IEEE Conference on Computer Vision and Pattern
  Recognition (CVPR)}.

\bibitem[Zhang et~al., 2016]{zhan:yin16}
Zhang, P., Goyal, Y., Summers{-}Stay, D., Batra, D., and Parikh, D. (2016).
\newblock Yin and yang: Balancing and answering binary visual questions.
\newblock In {\em Proceedings of CVPR}.

\bibitem[Zhou et~al., 2015]{zhou:simp15}
Zhou, B., Tian, Y., Suhkbaatar, S., Szlam, A., and Fergus, R. (2015).
\newblock Simple baseline for visual question answering.
\newblock Technical report, arXiv:1512.02167, 2015.

\end{thebibliography}
\bibliographystyle{apalike} 

\end{document}